\documentclass[aip,amsmath,amssymb,floatfix,reprint]{revtex4-1}

\usepackage{graphicx}
\usepackage{dcolumn}
\usepackage{bm}
\usepackage{lineno}

\DeclareMathOperator*{\argmin}{argmin}
\usepackage{subfigure}
\usepackage{float}
\usepackage{mathptmx}
\usepackage{newtxtext,newtxmath}
\usepackage{hyperref}
\usepackage{textcomp}

\begin{document}
\preprint{AIP/123-QED}
\title[Coarse-scale PDEs from fine-scale observations via machine learning]{Coarse-scale PDEs from fine-scale observations via machine learning}
\author{Seungjoon Lee}
\affiliation{Department of Chemical and Biomolecular Engineering, Johns Hopkins University}
\author{Mahdi Kooshkbaghi}%
\affiliation{Program in Applied and Computational Mathematics, Princeton University}
\author{Konstantinos Spiliotis}
\affiliation{Institute of Mathematics, University of Rostock}
\author{Constantinos I. Siettos}
\affiliation{Dipartimento di Matematica e Applicazioni ``Renato Caccioppoli", Universit\'a degli Studi di Napoli Federico II}
\author{Ioannis G. Kevrekidis}
\altaffiliation[Also at ]{Department of Applied Mathematics and Statistics; Department of Medicine, Johns Hopkins University}
\email{yannisk@jhu.edu}
\affiliation{Department of Chemical and Biomolecular Engineering, Johns Hopkins University}
\date{\today}
\begin{abstract}
Complex spatiotemporal dynamics of physicochemical processes are often modeled at a microscopic level (through e.g.~atomistic, agent-based or lattice models) based on first principles.
Some of these processes can also be successfully modeled at the macroscopic level using e.g.~partial differential equations (PDEs) describing the evolution of the right few macroscopic observables (e.g.~concentration and momentum fields). 
Deriving good macroscopic descriptions (the so-called ``closure problem'') is often a time-consuming process requiring deep understanding/intuition about the system of interest. 
Recent developments in data science provide alternative ways to effectively extract/learn accurate macroscopic descriptions approximating the underlying microscopic observations.
In this paper, we introduce a data-driven framework for the identification of unavailable coarse-scale PDEs from microscopic observations via machine learning algorithms. 
Specifically, using Gaussian Processes, Artificial Neural Networks, and/or Diffusion Maps, the proposed framework uncovers the relation between the relevant macroscopic space fields and their time evolution (the right-hand-side of the explicitly unavailable macroscopic PDE). 
Interestingly, several choices equally representative of the data can be discovered.
The framework will be illustrated through the data-driven discovery of macroscopic, concentration-level PDEs resulting from a fine-scale, Lattice Boltzmann level model of a reaction/transport process. 
Once the coarse evolution law is identified, it can be simulated to produce long-term macroscopic predictions.
Different features (pros as well as cons) of alternative machine learning
algorithms for performing this task (Gaussian Processes and Artificial Neural Networks), are presented and discussed.
\end{abstract}


\maketitle

\begin{quotation}
The behavior of microscopic complex systems is often described in terms of
effective, macroscopic governing equations, leading to simple and efficient prediction.
Yet, the discovery/derivation of such macroscopic governing equations generally relies on deep understanding and prior knowledge about the system,
as well as extensive and time-consuming mathematical justification.
Recent developments in data-driven computational approaches suggest  alternative ways towards uncovering useful coarse-scale governing equations directly from fine scale data. 
Interestingly, even deciding what the ``right'' coarse-scale variables are, may present a significant challenge. 
In this paper, we introduce and implement a framework for systematically extracting coarse-scale observables from microscopic/fine scale data and for discovering the underlying governing equations using machine learning techniques (e.g. Gaussian processes and artificial neural networks) enhanced by feature selection methods. 
Intrinsic representations of the coarse-scale behavior via manifold learning techniques (in particular, Diffusion Maps), generating alternative possible forms of the governing equations is also explored and discussed.
\end{quotation}
\section{Introduction}
The successful description of the spatiotemporal evolution of complex systems typically relies on detailed mathematical models operating at a fine scale (e.g.~molecular dynamics, agent-based, stochastic or lattice-based methods).
Such microscopic, first principles models, keeping track of the interactions between huge numbers microscopic level degrees of freedom, typically lead to prohibitive computational cost for large-scale spatiotemporal simulations.

To address this issue (and since we are typically interested in macro-scale features -pressure drops, reaction rates- rather than the position and velocity of each individual molecule), reduced, coarse-scale models are developed and used, leading to significant computational savings in large-scale spatiotemporal simulations~\cite{Noid13}.

Macroscopically, the fine scale processes may often be successfully modeled using partial differential equations (PDEs) in terms of the right macroscopic observables (``coarse variables": not molecules and their velocities, say, but rather pressure drops and momentum fields). 
Deriving the macroscopic PDE that effectively models the microscopic physics (the so-called ``closure problem'') requires, however, deep understanding/intuition about the complex system of interest and often extensive mathematical operations; the discovery of macroscopic governing equations is typically a difficult and time-consuming process.

To bypass the first principles discovery of a macroscopic PDE directly, several data-driven approaches provide ways to effectively determine good coarse observables and the approximate coarse-scale relations between them from simulation data.
In the early '90s researchers (including our group) employed artificial neural networks for system identification (both lumped and distributed)~\cite{Hudson90,Krischer93,Rico94,Anderson96,Gonzalez98}.
Projective time integration in dynamical systems~\cite{Kevrekidis03} and fluid dynamics~\cite{Sirisup05,Lee17D} also provides a good data-driven approximation of long-time prediction based not on closed-form equations, but rather on a ``black box" simulator.
Furthermore, the equation-free framework for designing fine scale computational experiments and systematically processing their results through ``coarse-time steppers'' has proven its usefulness / computational efficiency in analyzing macroscopic bifurcation diagrams~\cite{Siettos12,Theodoropoulos00}. 
The easy availability of  huge simulation data sets, and recent developments in the efficient implementation of machine learning algorithms, has made the revisiting of the identification of nonlinear dynamical systems from simulation time series an attractive -and successful- endeavor.
Working with observations at the macroscopic level, hidden macroscopic PDEs can be recovered directly by artificial neural networks~\cite{Gonzalez98}, (see also Ref.~\cite{Bar19}). Sparse identification of nonlinear dynamics (SINDy)~\cite{Rudy17} as well as Gaussian processes~\cite{Raissi17,Raissi18} have also been successfully used, resulting in {\em explicit} data-driven PDEs. All these approaches rely on macroscopic observations.

In this paper, we discuss the identification of unavailable coarse-scale PDEs {\em from fine scale observations} through a combination of machine learning and manifold learning algorithms. 
Specifically, using Gaussian Processes, Artificial Neural Networks, and/or Diffusion Maps, and starting with candidate coarse fields (e.g.~densities), our procedure extracts relevant macroscopic features (e.g.~coarse derivatives) from the data, and then uncovers the relations between these macroscopic
features and their time evolution (the right-hand-side of the explicitly unavailable macroscopic PDE). 

To effectively reduce the input data domain, we employ two feature selection methods: (1) a sensitivity analysis via Automatic Relevance Determination (ARD)~\cite{Qi04,Rasmussen06,Wipf08} in Gaussian processes and (2) a manifold learning technique, Diffusion Maps~\cite{Holiday19}.
Having selected the relevant macro features in terms of which the evolution can be modelled, we employ two machine learning algorithms to approximate a ``good'' right-hand-side of the underlying PDEs: (1) Gaussian process regression and (2) artificial neural networks.

Our framework is illustrated through the data-driven discovery of the macroscopic, concentration-level PDE resulting from a fine-scale, Lattice Boltzmann (LB) model of a reaction/transport process (the FitzHugh-Nagumo process in one spatial dimension). 
Long-term macroscopic prediction is enabled by numerical simulation of the coarse-scale PDE {\em identified from the Lattice-Boltzmann data}.
Different possible feature combinations (leading to different realizations of the same evolution) will also be discussed.

The remainder of the paper is organized as follows: In section~\ref{sec:framework}, we present an overview of our proposed framework and briefly review theoretical concepts of Gaussian process regression, artificial neural networks, and Diffusion Maps.
Two methods for feature selection are also presented.
In section~\ref{sec:simulation}, we describe two simulators at different scales: (1) the FitzHugh-Nagumo model at the macro-scale and (2) its Lattice Boltzmann realization at the micro-scale.
In section~\ref{sec:result}, we demonstrate the effectiveness of our proposed framework and discuss the advantages and challenges of different feature selection methods and regression models for performing this task.
In section \ref{sec:conclusion}, we summarize our results and discuss open issues for further development of the data-driven discovery of the underlying coarse PDE from microscopic observations.
\section{Framework for recovering a coarse-scale PDE via machine learning}
\label{sec:framework}
\subsection{Overview}
\begin{figure}[!htp]
  \centering
  \includegraphics[width=0.45\textwidth]{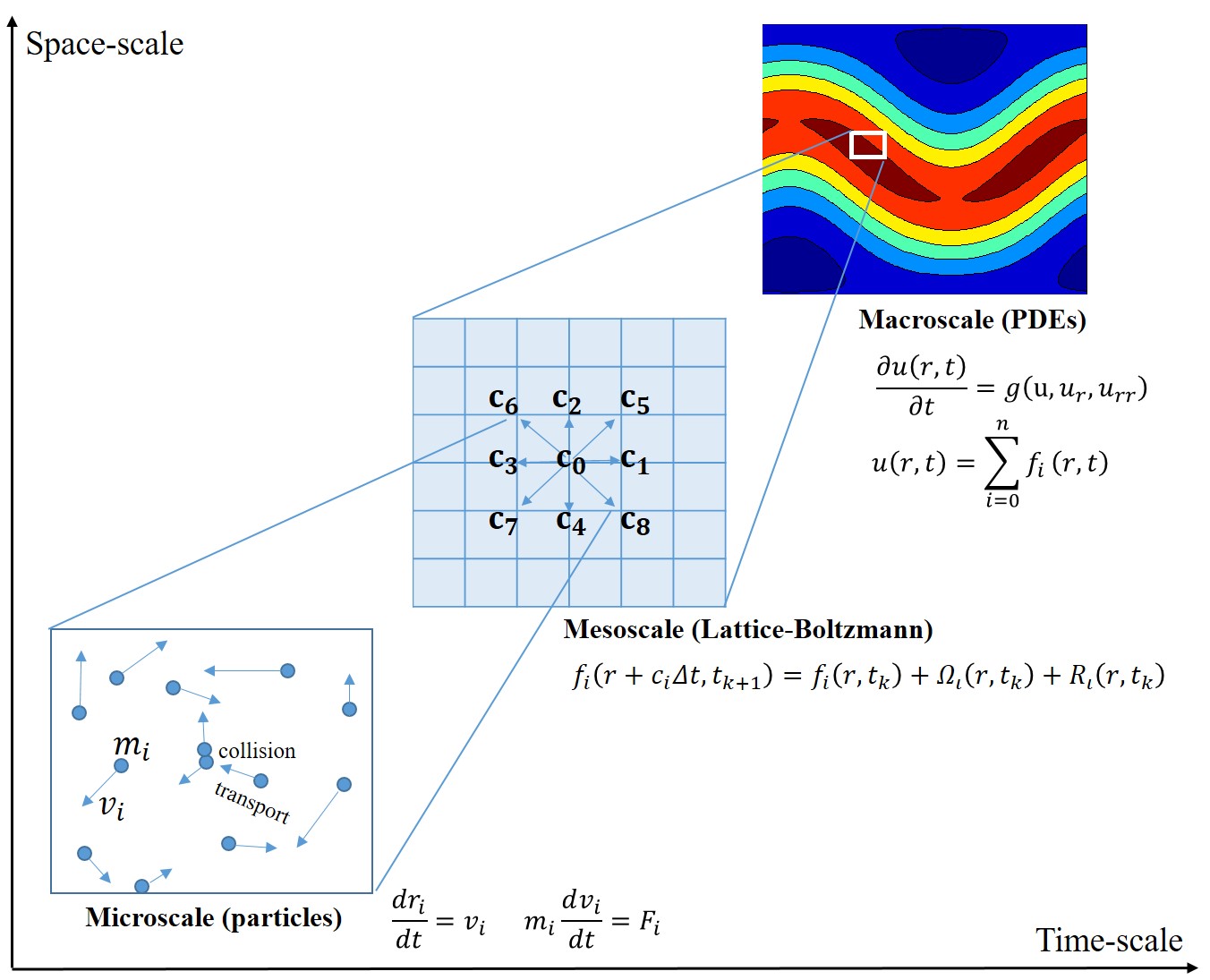}
  \caption{Schematic illustration of the extraction of coarse-scale observables $u$ from microscopic observations.
  Through a Lattice Boltzmann model (here, D2Q9), we obtain particle distribution functions ($f_i$)  on a given lattice.
  Using the zeroth moment field of the particle distribution function at the grid point $x_n$, we extract the coarse observable $u$ (in this paper, we have two coarse observables, $u$ and $v$, which represent the density of the activator and the inhibitor, respectively).
  }
  \label{fig:LBM}
\end{figure}
\begin{figure*}[!htp]
  \centering
  \includegraphics[scale=0.48]{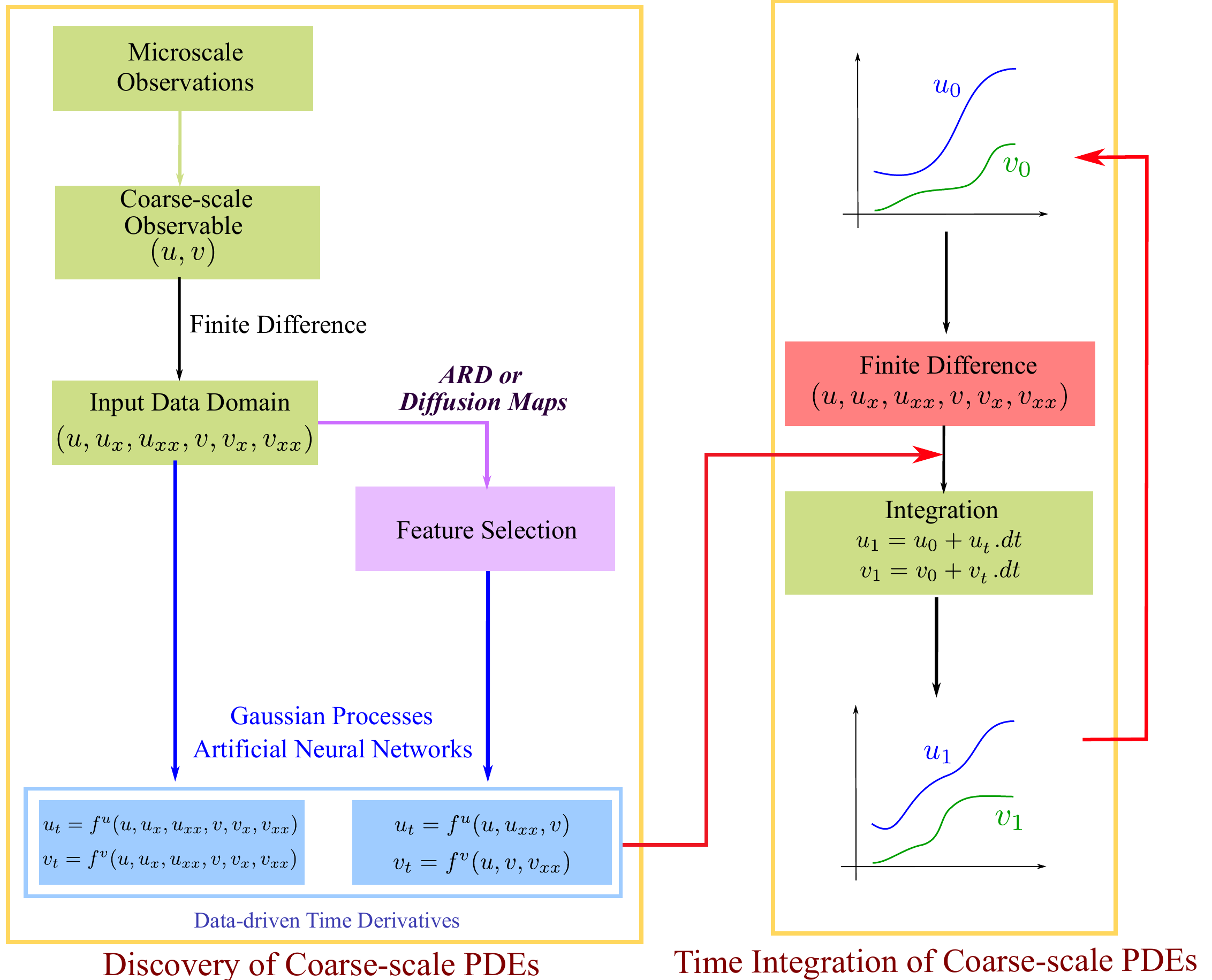}
  \caption{Workflow for uncovering coarse-scale PDEs.
  First, we compute macroscopic variables $u$ and $v$ from the Lattice Boltzmann simulation data (see equation~\eqref{eqn:concentration} and figure~\ref{fig:LBM}) and estimate their spatial derivatives (e.g. by finite difference schemes on the lattice).
  After that, we employ machine learning algorithms (here, Gaussian process regression or artificial neural networks) to identify ``proper'' time derivatives $u_t$ and $v_t$ from an original input data domain directly (no feature selection among several spatial derivatives) or from a reduced input data domain (feature selection among several spatial derivatives) using ARD in Gaussian processes or Diffusion Maps.
  We then simulate the identified coarse-scale PDE for given coarse initial conditions ($\mathbf{u_0},\mathbf{v_0}$).
  }
  \label{fig:frame}
\end{figure*}
The workflow of our framework for recovering hidden coarse-scale PDEs from microscopic observations is schematically shown in figures~\ref{fig:LBM} and~\ref{fig:frame}.
Specifically, this framework consists of two subsections: (1) computing coarse-scale observables and (2) identifying coarse-scale PDEs and then numerically simulating them.

To clarify the algorithm, consider a single field (the activator $u$; later in this paper we will use two densities, $u$ and $v$, for the activator and the inhibitor, respectively).
As shown in figure~\ref{fig:LBM}, we compute the coarse-scale observable (here the $u$ concentration field) through the zeroth moment of the microscopic LB simulation (averaging the particle distribution functions ($f_i$) on a given lattice point, see section~\ref{sec:lbm} for more details).

Given the coarse-scale observable  we estimate its time-derivative and several of its spatial derivatives (e.g.~$u_t$, $u_x$, and $u_{xx}$), typically using finite difference schemes in time and space as
necessary.
A PDE of the form $u_t=L(u)= F(u, u_x, u_{xx}, \cdots)$ is a relation between the time-derivative and a number of spatial derivatives; this relation holds at every moment in time and every point in space. 
For a simple reaction diffusion equation, say $u_t=u_{xx} - u$, the data triplets $(u, u_t, u_{xx})$ will in general lie on a two-dimensional manifold in three-dimensional space, since $u_t$ is a function of $u$ and $u_{xx}$. 
Knowing that this manifold is two-dimensional suggests (in the spirit of the the Whitney and Takens embedding theorems~\cite{Whitney36, Takens81}) that any five generic observables suffice to create an embedding - and thus learn $u_t$, a function on the manifold, as a function of these five observables.
One might choose, for example, as observables the values of $u$ at any five spatial points at a given time moment, possibly the five points used in a finite difference stencil for estimating spatial derivatives.
In the study of time series through delay embeddings one uses observations on a temporal stencil; it is interesting that here one might use observations on a spatial stencil - encoding information analogous to spatial derivatives (see Ref.~\cite{Bar19}).
Motivated by this perspective, in order to learn the time derivative $u_t$, we use an original input data domain including several (say, all up to some order) spatial derivatives. We also consider the selection of a reduced input data domain via two feature selection methods: (1) a sensitivity analysis by automatic relevance determination (ARD) in Gaussian processes~\cite{Williams96,Qi04,Wipf08} and (2) a manifold learning approach, Diffusion Maps~\cite{Coifman05,Coifman06}, with a regression loss (see section~\ref{sec:fs} in more details).
Then, we consider two different machine learning methods (Gaussian process regression and artificial neural networks) to learn $u_t$ based on the selected feature input data domain.

After training, simulation of the learned coarse-scale PDE given a coarse initial condition $u_0(x,t), v_0(x,t)$ can proceed with any acceptable discretization scheme in time and space (from simple finite differences to, say, high order spectral or finite element methods). 
\subsection{Gaussian process regression}
One of the two approaches we employ to extract dominant features and uncover the RHS of coarse-scale PDEs is Gaussian process regression. 
In Gaussian processes, to represent a probability distribution over target functions (here, the time derivative), we assume that our observations are a set of random variables whose finite collections have a multivariate Gaussian distribution with an {\em unknown} mean (usually set to zero) and an {\em unknown} covariance matrix $K$.
This covariance matrix is commonly formulated by a Euclidean distance-based kernel function $\kappa$ in the input space, whose hyperparameters are optimized by training data.
Here, we employ a radial basis kernel function (RBF), which is the \emph{de facto} default kernel function in Gaussian process regression, with ARD~\cite{Rasmussen06}.
\begin{equation}
K_{ij}=\kappa(\mathbf{x_i},\mathbf{x_j};\theta) = \theta_0\exp \left( -\frac{1}{2} \sum_{l=1}^k \frac{(x_{i,l} - x_{j,l})^2}{\theta_l
} \right).
\label{eqn:kernel}
\end{equation} 
where $\theta = [\theta_0, \dots, \theta_k]^T$ is a $k+1$ dimensional vector of hyperparameters and $k$ is the number of dimensions of the input data domain. 
The optimal hyperparameter set $\theta^*$ can be obtained by minimizing a negative log marginal likelihood with the training data set $\{\mathbf{x},\mathbf{y}\}$:
\begin{eqnarray}
\label{eqn:opt}
\theta^* &=& \argmin_{\theta} \; \{ -\log p(\mathbf{y}|\mathbf{x},\theta)\}  \\
&=& \frac{1}{2}\mathbf{y}^T(K+\sigma^2I)^{-1}\mathbf{y} + \frac{1}{2}\log|(K+\sigma^2I)| + \frac{N}{2}\log  2\pi \nonumber
\end{eqnarray}
where $N$ is the number of training data points, $\sigma^2$ and $I$ represent the variance of the (Gaussian) observation noise and an $N \times N$ identity matrix, respectively.

To find the Gaussian distribution of the function values (here the time derivative) at test data points, we represent the multivariate Gaussian distribution with the covariance matrix constructed by equation~\eqref{eqn:kernel} as 
\begin{equation}
\begin{bmatrix} 
\mathbf{y} \\
\mathbf{y^*} 
\end{bmatrix}
=
N \left( 
\mathbf{0},
\begin{bmatrix} 
K + \sigma^2 I &  K_*\\
K_*^T & K_{**} 
\end{bmatrix}
\right),
\end{equation}
where $\mathbf{y^*}$ is a predictive distribution for test data $\mathbf{x^*}$, $K_*$ represents a covariance matrix between training and test data while $K_{**}$ represents a covariance matrix between test data.

Finally, we represent a Gaussian distribution for time derivatives at the test point in terms of a predictive mean and its variance, through  conditioning a multivariate Gaussian distribution as
\begin{equation}
\mathbf{\bar{y}^*} = K_*(K+\sigma^2I)^{-1}\mathbf{y},
\end{equation}
\begin{equation}
K(\mathbf{y^*}) = K_{**} - K_*^T(K+\sigma^2I)^{-1}K_*,
\end{equation}
and we assign the predictive mean ($\mathbf{\bar{y}^*}$) as the estimated time derivative for the corresponding data point. 
\subsection{Artificial neural networks}
\label{sec:ANN}
Next, we consider (artificial, possibly deep) neural networks (ANN or NN or DNN) for identifying the RHS of coarse-scale PDEs.
Generally, neural networks consist of an input layer, one or more hidden layers, and an output layer, all comprised of several computational neurons, typically fully connected by weights ($\omega$), biases ($b$), and an activation function ($\psi(\cdot)$).
Macroscopic observables and their spatial derivatives are assigned at the input layer, while the corresponding time derivative is obtained at the  output layer (here we are considering only first order PDEs in time; higher order equations, like the wave equation, involving second derivatives in time can also be accounted for within the framework).
In (feed-forward) neural networks, a universal approximation theorem~\cite{Cybenko89} guarantees that for a single hidden layer with (sufficient) finite number of neurons, an approximate realization $\tilde{y}$ of the target function, $y$ can be found. 
Here, approximation implies that the target and learned functions are sufficiently close in an appropriately chosen norm ($\forall \delta >0: \Vert y - \tilde{y} \Vert < \delta$). 
The approximate form of the target function obtained through the feedforward neural net can be written as
\begin{equation}
\tilde{y}(\mathbf{x}) = \sum_{i=1}^N \psi \left(\mathbf{\omega}_i^{T}\mathbf{x} + b_i \right).
\end{equation}
\noindent
The root-mean-square error cost function 
\begin{equation}
E_D = \frac{1}{N} \sum_{j=1}^N(y_j - \tilde{y}(x_j))^2,
\end{equation}
typically measures the goodness of the approximation.

In order to obtain a \textit{generalizable} network, with
good performance on the test data set as well as on the training data set (e.g.~preventing overfitting), several regularization approaches have been proposed, mostly relying on modifications of the cost function.
Foresee and Hagan~\cite{Foresee97} showed that modifying the cost function by adding the regularization term $E_\omega=\Sigma_{j=1}^{N_\omega}\omega_j^2$,  results in a network that will maximize the posterior probability based on Bayes' rule. 
We thus trained our network based on a total cost function of the form:
\begin{equation}
    E_{total} = \beta_1 E_D + \beta_2 E_\omega,
\end{equation}
in which $\beta_1$ and $\beta_2$ are network tuning parameters.
Here, we employ Bayesian regularized back-propagation for training, which updates weight and bias values through Levenberg-Marquardt optimization~\cite{Hagan94}; we expect that, for our data, comparable results would be obtained through other modern regularization/optimization algorithms.
\subsection{Diffusion Maps}
\label{sec:dmap}
Diffusion Maps (DMAP) have successfully been employed for dimensionality reduction and nonlinear manifold learning~\cite{Coifman05,Coifman06,Nadler06,Nadler08}.
The Diffusion Maps algorithm can guarantee, for data lying on a smooth manifold -and at the limit of infinite data- that the eigenvectors of the large normalized kernel matrices constructed from the data converge to the eigenfunctions of the Laplace-Beltrami operator on the manifold on which the data lie.
These eigenfunctions can also provide nonlinear parametrizations (i.e.~sets of coordinates) for such a (Riemannian) manifolds.
To approximate the Laplace-Beltrami operator from scattered data points on the manifold, a normalized diffusion kernel matrix between observation (data) points is commonly used:
\begin{equation}
    \mathbf{W_{ij}} = \exp \left( -\frac{\Vert \mathbf{y_i}-\mathbf{y_j}\Vert_{2}^2}{\varepsilon}\right),
    \end{equation}
where $\mathbf{y_i}$ are real-valued observations and $\varepsilon$ is the kernel width.
After that, one obtains a normalized matrix $\mathbf{W}^{(\alpha)}$ by
\begin{equation}
\mathbf{W}^{(\alpha)} = \mathbf{D}^{-\alpha}\mathbf{W}\mathbf{D}^{-\alpha},
\end{equation}
where $\mathbf{D}$ is a diagonal matrix whose $\mathrm{i^{th}}$ entry is the sum of corresponding row of $W$.
Here, $\alpha\in\{0,1\}$ is a tuning parameter: 
$\alpha=0$ corresponds to the classical normalized graph Laplacian~\cite{Belkin02,Belkin03} while $\alpha=1$, which takes into account the local data density, yields the Laplace-Beltrami operator~\cite{Coifman06}; in this paper, we set $\alpha=1$.
Then, $\tilde{\mathbf{W}}$ is calculated simply as:
\begin{equation}
\tilde{\mathbf{W}} = \tilde{\mathbf{D}}^{-1}\mathbf{W}^{(\alpha)},
\end{equation}
where $\tilde{\mathbf{D}}$ is a diagonal matrix whose $\mathrm{i^{th}}$ entry is the sum of corresponding row of $\mathbf{W}^{(\alpha)}$.

Finally, an embedding of the manifold is constructed by the first few (say $m$) nontrivial eigenvectors of $\mathbf{\tilde{W}}$,
\begin{equation}
\mathbf{y_i} \mapsto \left(\lambda_1^t\phi_{1,i},\dots,\lambda_m^t\phi_{m,i}\right),\;\; i=1,\dots,n,
\end{equation}
where $t$ corresponds to the number of diffusion steps (here $t=0$) with descending ordered eigenvalues $\lambda_i$.

Instead of using the Euclidean distance between the data points in the Diffusion Map algorithm, it has recently been proposed to use a different, kernel-based similarity metric\cite{Bittracher19}; the associated kernel-based embeddings allow for control of the distortion of the resulting embedding manifold; we will return to this and its possible implications in our Conclusions (Section~\ref{sec:conclusion}).

\subsection{Feature selection}
\label{sec:FS}
Describing the coarse-scale spatiotemporal dynamics in the form of a PDE, involves learning the local field time-derivatives
as a function of a few, relevant local field spatial derivatives. 
Starting with a ``full" input data domain consisting of all local field values as well as all their coarse-scale spatial derivatives (up to some order),
we must extract the few ``relevant" spatial derivatives as dominant features of this input data domain.
Such feature selection will typically reduce the dimensionality of the input data domain.
Among various feature selection methods, we employ two algorithms based on (1) sensitivity analysis via ARD in Gaussian processes~\cite{Williams96,Qi04,Wipf08} and (2) manifold parametrization through output-informed Diffusion Maps~\cite{Holiday19}.

First, we employ sensitivity analysis via automatic relevance determination (ARD) in Gaussian processes, which effectively reduces the number of input data dimensions.
In Gaussian processes, we obtain the optimal hyperparameter set $\theta^*$ by minimizing a negative log marginal likelihood (see equation~\eqref{eqn:opt}).
ARD assigns a different hyperparameter $\theta_i$ for each input dimension $d_i$.
As can be seen in equation~\eqref{eqn:kernel}, a large value of $\theta_i$ nullifies the difference between target function values along the $d_i$ dimension, allowing us to designate this dimension as ``insignificant". 
Practically, we select the input dimensions with relatively small $\theta_j$ to build a reduced input data domain, which can still successfully represent the approximation of right-hand-side on the underlying PDEs. 
\begin{figure*}[!htp]
  \centering
  \includegraphics[width=0.8\textwidth]{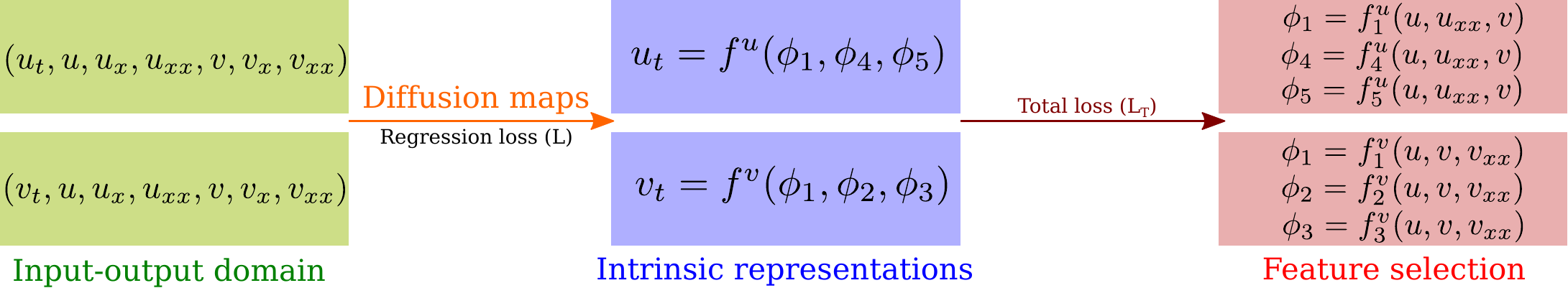}
  \caption{Input feature selection via output-informed Diffusion Maps.
  Diffusion Maps provide intrinsic coordinatization of the output (the time derivatives) from combined input-output 
  data.
  Guided by a regression loss ($L$), we find a low-dimensional intrinsic embedding space in which $u_t$ (and $v_t$) can be represented as a function of just a few intrinsic diffusion
  map coordinates.
  After that, we search and find minimal subsets of the input data domain that can parametrize the selected intrinsic coordinates (e.g. $\phi_1,\phi_4,\phi_5$) 
  as quantified by a small total regression loss (see equation~\eqref{eqn:tloss}).
  }
  \label{fig:fdmap}
\end{figure*}

Alternatively, we employ a manifold learning technique to find the intrinsic representation of the coarse-scale PDE, and then examine the relation between these intrinsic coordinates and given input features (spatial field derivatives).
Specifically, Diffusion Maps will provide an intrinsic parametrization of the combined input-output data domain (here, $\{u_t,u,u_x,u_{xx},v,v_x,v_{xx} \}$ for $u$ and $\{v_t,u,u_x,u_{xx},v,v_x,v_{xx} \}$ for $v$).
Selecting leading intrinsic coordinates, we can then find the lowest-dimensional embedding space for the PDE manifold (the manifold embodying $u_t$ and $v_t$ as a function of the embedding intrinsic coordinates.) 
We then test several combinations of subsets of the input domain coordinates (spatial derivatives) as to their ability to parametrize the discovered intrinsic embedding coordinates.
Each such set of such inputs, that successfully parametrize the intrinsic embedding coordinates, provides us a new possibility of learning a PDE formulation that describes the spatiotemporal dynamics of our observation data set.

In principle, any subset of intrinsic coordinates that successfully parametrizes the manifold can be used to learn functions on the manifold, and in particular $u_t$ and $v_t$.
The success of any particular subset of leading intrinsic coordinates in so describing $u_t$ and $v_t$ is confirmed through regression, via a mean-squared-error loss ($L$).

Next, we investigate which set of features of the input domain (which sets of spatial derivatives) can be best used to parametrize the intrinsic embedding (and thus learn the PDE right-hand-side).
One can find the subset of features from a user-defined dictionary (here spatial derivatives) to parametrize the intrinsic embedding coordinates through a linear Group LASSO~\cite{Meila18}.
In this paper, we examine several combinations of input domain variables, and find subsets that can minimally parametrize the intrinsic embedding; this is quantified through a total regression loss ($L_T$) based on a mean-squared-error as
\begin{equation}
\label{eqn:tloss}
    L_T = \left(\sum_{j=1}^{d}L_{\phi_j}^2\right)^{\frac{1}{2}},
\end{equation}
where $L_{\phi_j}$ represents a regression loss for representing the intrinsic coordinate $\phi_j$ using selected input features and $d$ represents the number of intrinsic coordinates we chose.

ARD for Gaussian processes suggests the ``best" input domain subset in terms of which we will try and predict $u_t$ and $v_t$.
In the manifold learning context, we may find several different input subsets capable of parametrizing the manifold on which the observed behavior lies. 
Different minimal parametrizing subsets will lead to different (but, in principle, on the data, equivalent) right-hand-sides for the PDE evolution. One expects that some of them will be ``better conditioned" (have better Lipschitz constants) than others. 
\section{Different scale simulators for one-dimensional reaction-diffusion systems}
\label{sec:simulation}
\subsection{Macro-scale simulator: FitzHugh-Nagumo model}
\label{sec:fhn}
\begin{figure}[!htp]
  \centering
      \subfigure[~$u$ by LBM]{
  \includegraphics[scale=0.125]{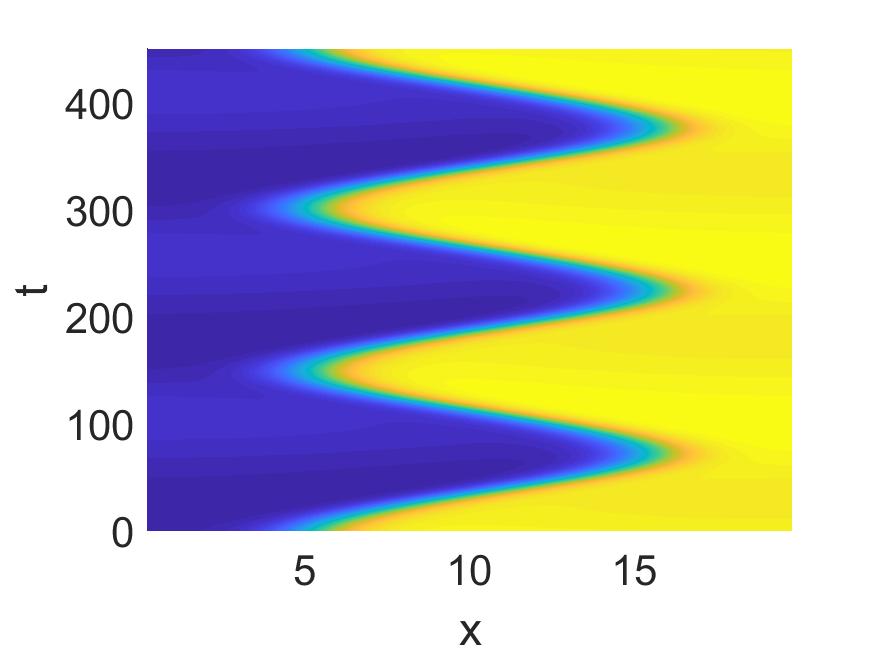}
  } 
    \subfigure[~$u$ by FHN]{
  \includegraphics[scale=0.125]{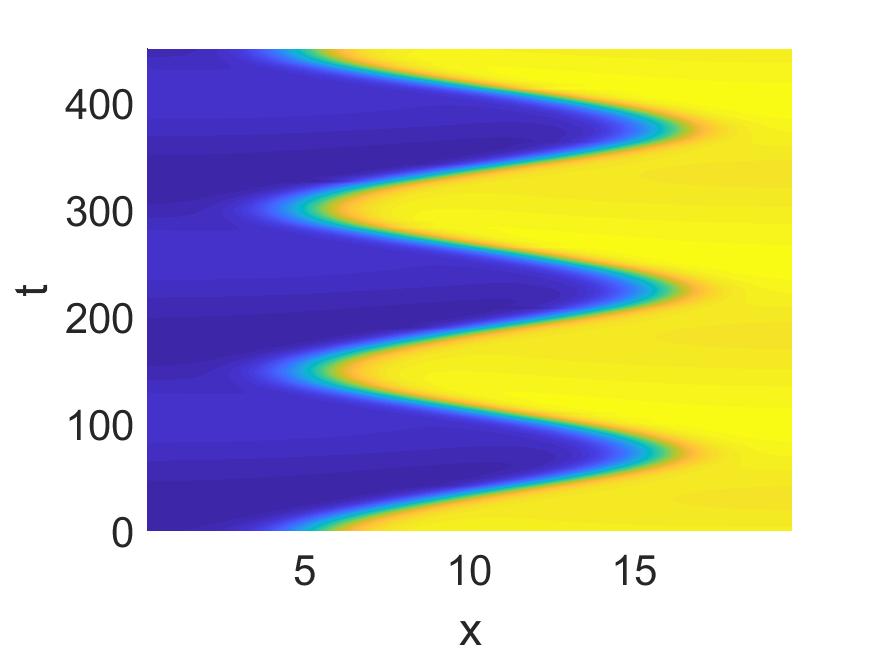}
  }
    \subfigure[~$v$ by LBM]{
  \includegraphics[scale=0.125]{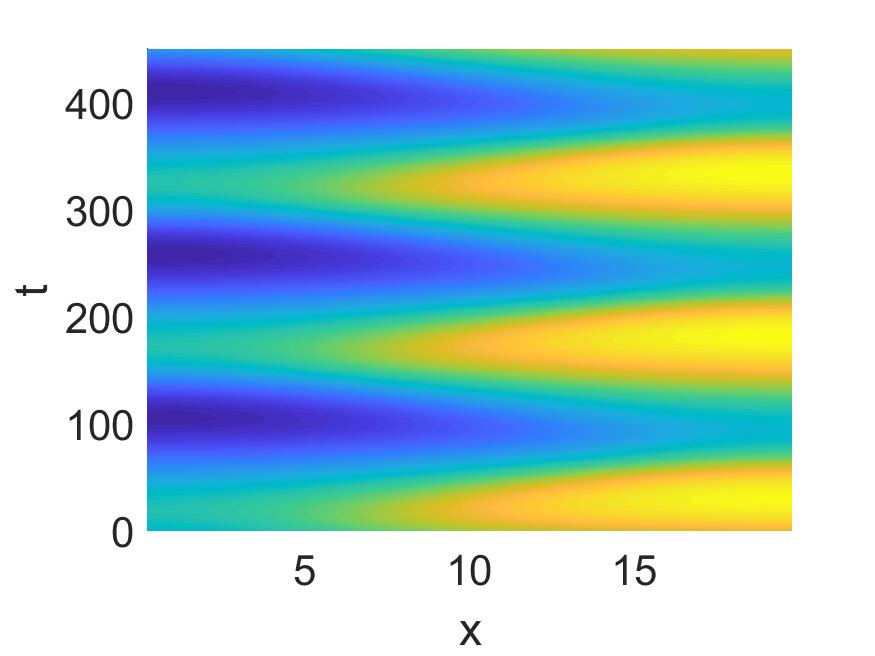}
  }
     \subfigure[~$v$ by FHN]{
  \includegraphics[scale=0.125]{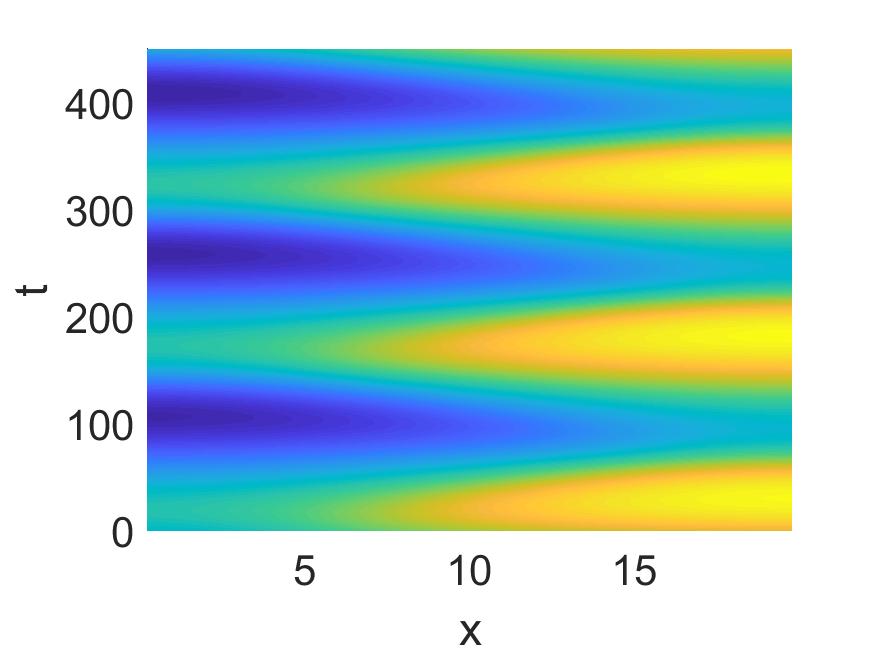}
  }  
    \subfigure[~Absolute difference for $u$]{
  \includegraphics[scale=0.125]{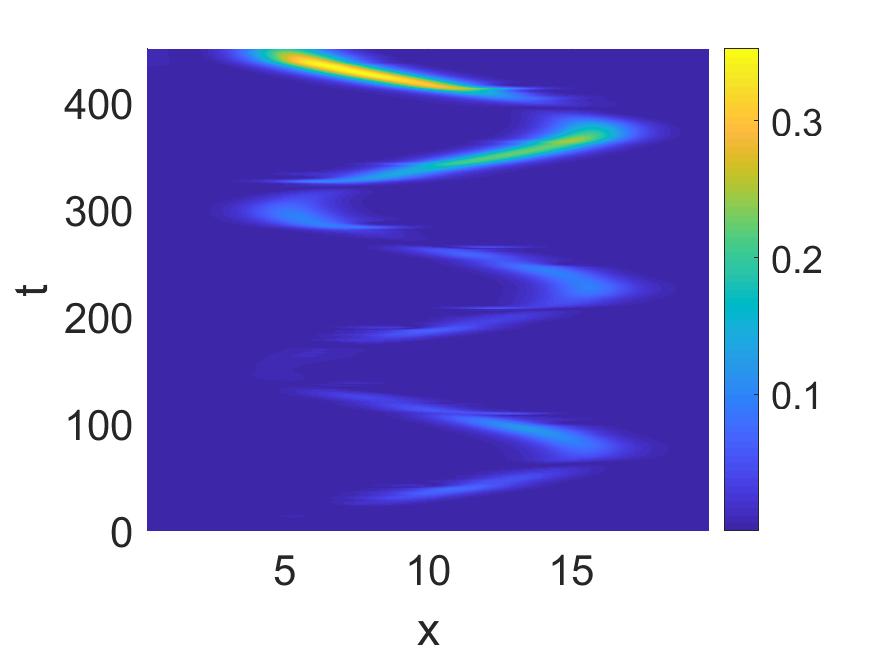}
  } 
     \subfigure[~Absolute difference for $v$]{
  \includegraphics[scale=0.125]{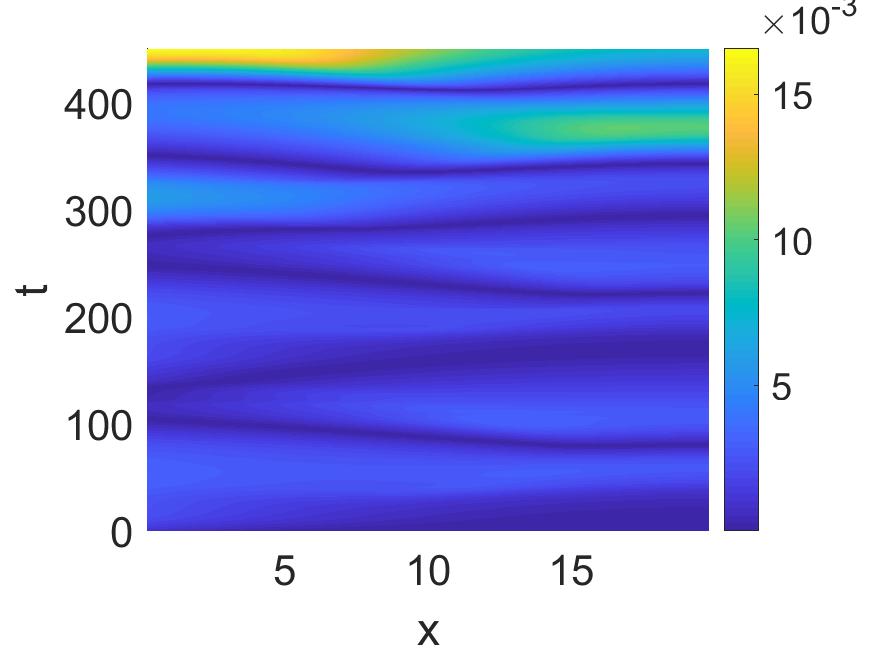}
  }    
  \caption{Spatiotemporal behavior of $u$ and $v$ simulated by the Lattice-Boltzmann model and by the FitzHugh-Nagumo PDE.
  (a) and (c): $u$ and $v$ from the Lattice Boltzmann model (LBM). (b) and (d): $u$ and $v$ from the FitzHugh-Nagumo PDE.
  (e) and (f): Normalized absolute difference between the simulations of the two models.
  }
  \label{fig:origLBM}
\end{figure}
To describe a one-dimensional reaction-diffusion system that involves an activator $u$ and an inhibitor $v$, 
the FitzHugh-Nagumo model consists of two coupled reaction-diffusion partial differential equations:
\begin{equation}
\label{eqn:fhn}
\begin{aligned}
    \frac{\partial u}{\partial t} &= D^{u}\frac{\partial^2u}{\partial x^2} + u -u^3 - v,\\
    \frac{\partial v}{\partial t} &= D^{v}\frac{\partial^2v}{\partial x^2} + \epsilon(u -a_1v - a_0),
\end{aligned}
\end{equation}
where $a_1$ and $a_0$ are model parameters, $\epsilon$ represents a kinetic bifurcation parameter, and $D^{u}$ and $D^{v}$ represent diffusion coefficients for $u$ and $v$, respectively.
Here, we set these parameters to $a_1=2$, $a_0=-0.03$, $\epsilon=0.01$,  $D^{u}=1$, and $D^{v}=4$~\cite{Theodoropoulos00}.
We discretize a spatial domain on $[0, 20]$ with $\Delta x = 0.2$ and a time domain on $[0, 450]$ with $\Delta t=0.001$, respectively.
We impose homogeneous Neumann boundary conditions at both boundaries and solve these equations (for various initial conditions) numerically via the finite element method using the COMSOL Multiphysics\textregistered  ~software \cite{COMSOL}.
\subsection{Micro-scale simulator: the Lattice Boltzmann model}
\label{sec:lbm}
We also introduce a Lattice Boltzmann model (LBM)~\cite{Chen98,Succi01}, which can be thought of as a mesoscopic numerical scheme for describing spatiotemporal dynamics using finite-difference-type discretizations of Boltzmann-BGK equations~\cite{Bhatnagar54}, retaining certain advantages of microscopic particle models.
In this paper, the Lattice Boltzmann model is our fine scale ``microscopic simulator" and its results are considered to be ``the truth" from which the coarse-scale PDE will be learned. 

The time evolution of the particle distribution function on a given lattice can be described by 
\begin{equation}
\label{eqn:lbm}
    f^{l}_i(x_{j+i},t_{k+1}) = f^l_i(x_j,t_k) + \Omega^l_i(x_j,t_k) + R^l_i(x_j,t_k) \;\;\; l\in\{u, v\},
\end{equation}
where a superscript $l$ indicates the activator $u$ and the inhibitor $v$, and $\Omega^l_i$ represents a collision term defined by Bhatnagar-Gross-Krook (BGK)~\cite{Bhatnagar54}:
\begin{equation}
\Omega^l_i(x_j,t_k) = -\omega^l(f^l_i(x_j,t_k)-f_i^{l,eq}(x_j,t_k)),
\end{equation}
where $\omega^l$ represents a BGK relaxation coefficient defined as~\cite{Qian95}
\begin{equation}
    \omega^l = \frac{2}{1+3D^l\frac{\Delta t}{\Delta x^2}}.
\end{equation}

To compute our coarse-scale observables $u$ and $v$, we employ the D1Q3 model, which uses three distribution functions on the one-dimensional lattice as $(f^l_{-1}, f^l_0, f^l_{1})$ for each density (totalling 6 distribution functions).
Through the zeroth moment (in the velocity directions) of the overall distribution function, finally we compute the coarse-scale observable $u$ and $v$ as 
\begin{equation}
\label{eqn:concentration}
\begin{aligned}
 u(x_j,t_k) &= \sum_{i=-1}^{1}f^u_i(x_j,t_k),\\
 v(x_j,t_k) &= \sum_{i=-1}^{1}f^v_i(x_j,t_k).
\end{aligned}
\end{equation}

Based on spatially uniform local diffusion equilibrium, for which $f_i^{eq}$ is homogeneous in all velocity directions, the weights are chosen all equal to 1/3:
\begin{equation}
\begin{aligned}
    f_i^{u,eq}(x_j,t_k) &= \frac{1}{3}u(x_i,t_k),\\
    f_i^{v,eq}(x_j,t_k) &= \frac{1}{3}v(x_i,t_k).
\end{aligned}    
\end{equation}
Thus, the reaction terms $R^l_i$ in equation~\eqref{eqn:lbm} are modeled by
\begin{equation}
\label{eqn:reaction}
\begin{aligned}
    R_i^{u}(x_j,t_k) &= \frac{1}{3}\Delta t(u(x_j,t_k)-u(x_j,t_k)^3-v(x_j,t_k)),\\
    R_i^{v}(x_j,t_k) &= \frac{1}{3}\Delta t \epsilon (u(x_j,t_k)-a_1 v(x_j,t_k)^3-a_0).
\end{aligned}
\end{equation}
All model parameters ($a_0, a_1, \epsilon, D^u, D^v$) are the same as the FHN PDE.
The corresponding spatiotemporal behavior of these coarse observables $u$ and $v$ is shown in figures~\ref{fig:origLBM}(a) and (c) while the FHN PDE simulation for the same coarse initial conditions is shown in 
figures~\ref{fig:origLBM}(b) and (d).
\section{Results}
\label{sec:result}
\subsection{Learning without feature selection}
We begin by considering our proposed framework without feature selection, so as to later contrast with the results including feature selection.
\begin{figure}[!htp]
\centering
\begin{tabular}{cc}
\includegraphics[width=0.233\textwidth]{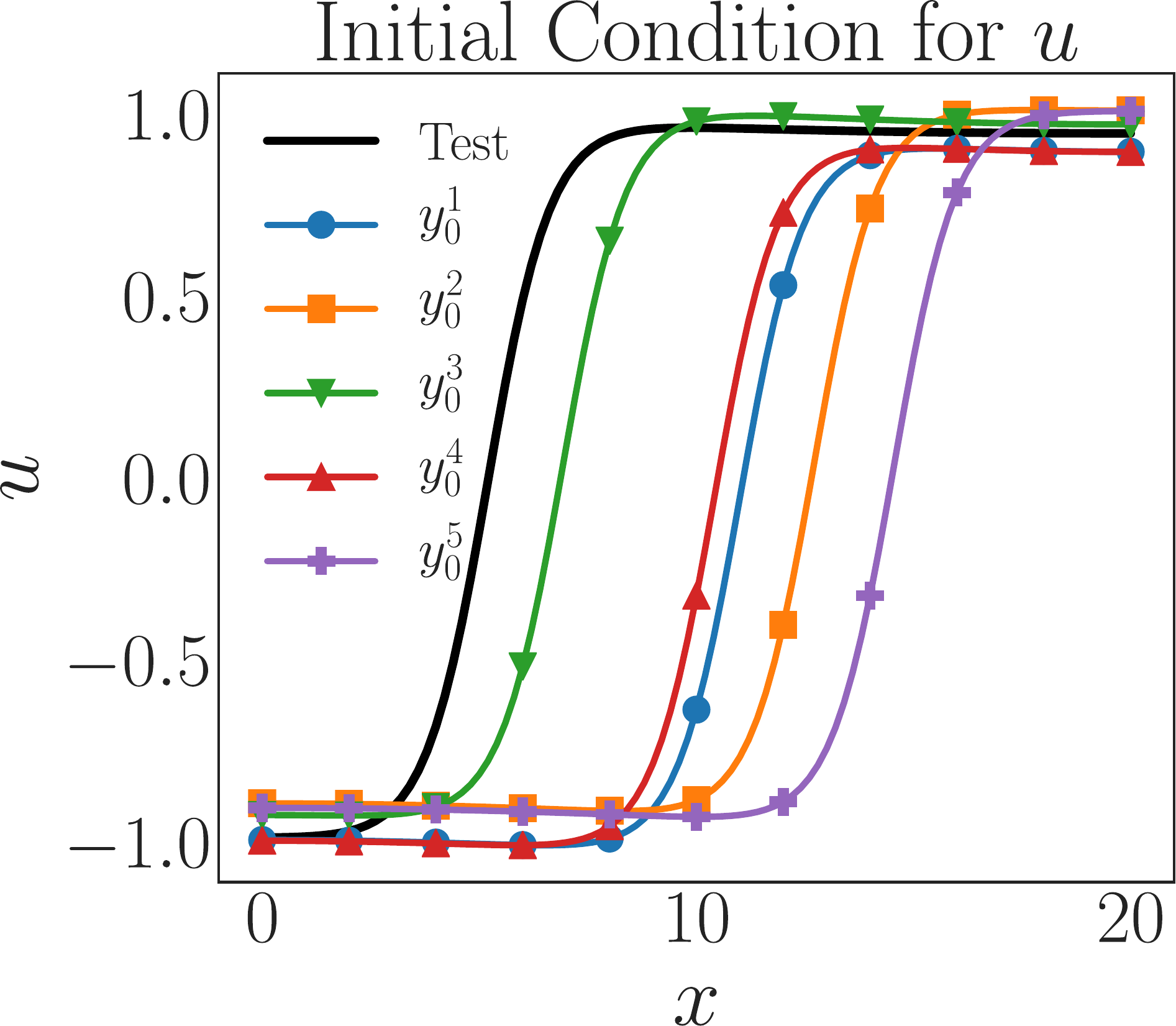} & 
\includegraphics[width=0.233\textwidth]{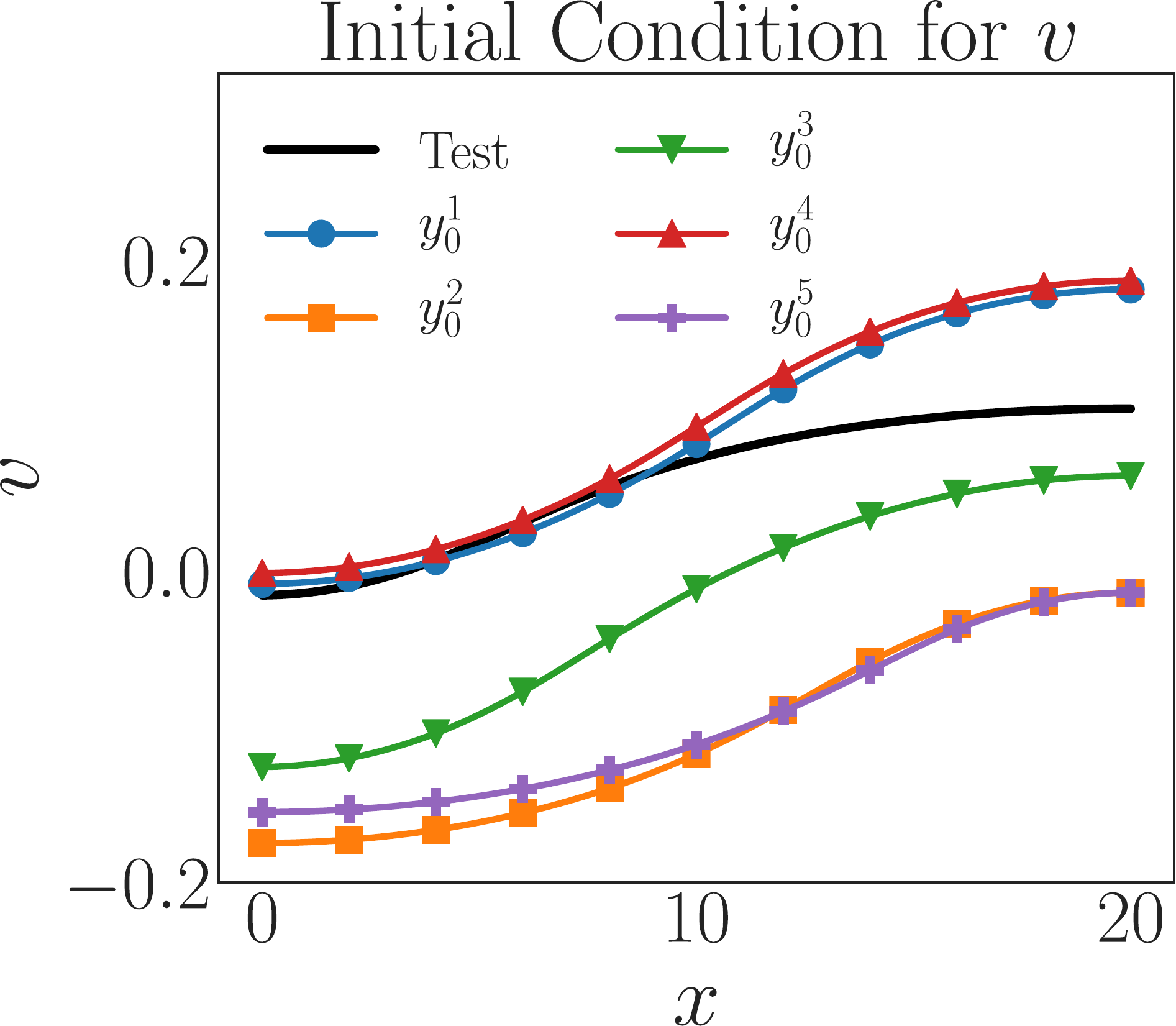} \\
(a) &
(b)
\end{tabular}
  \caption{Five different coarse initial conditions for training and a test coarse initial condition (colored in black). (a) Coarse initial conditions for $u$. (b) Coarse initial conditions for $v$. Five initial conditions are randomly chosen near the stable periodic solution.}
  \label{fig:ini}
\end{figure}
%
The data come from the fine-scale Lattice Boltzmann simulation. For the parameter values selected, the long-term dynamics of the LB simulation lie, for all practical purposes, on a stable time-periodic solution. 
To predict the coarse time derivatives $u_t$ and $v_t$, we collect training data from five different initial conditions near this stable periodic solution (see figure~\ref{fig:ini}) with the following LB spatiotemporal discretization -- in space, 99 discretized points on $[0.2, 19.8]$ with $dx = 0.2$; and in time, 451 discretized points on $[0, 450]$ with $dt=1$ for each initial condition.
Since our data come from the fine scale LB code, we need to initialize at the fine, LB scale of particle distribution functions (and not just of the concentrations $u$ and $v$). 
To initialize the particle distribution functions in the Lattice Boltzmann model we apply the equal weights rule, $1/3$ for $f_{-1}$, $f_{0}$, and $f_{1}$, motivated by near-equilibrium considerations.
We do expect that such initialization features will soon be ``forgotten" as higher distribution moments become quickly slaved to the lower (here the zeroth) moments
(see for example~\cite{Van05}).
To ensure that our results are not affected by the initialization details, we only start collecting training data after relaxation by short time simulation (here, 2000 time steps with $\Delta t=0.001$ or $t=2$), see appendix~\ref{sec:heal}.
We estimate the local coarse fields and their (several) spatial and temporal derivatives through finite differences, and then apply machine learning algorithms (here Gaussian processes as well as neural networks) to learn the time derivatives of the activator $u_t$ and the inhibitor $v_t$ using as input variables the local $u$, $v$ and all their spatial derivatives up to and including order two ($u, u_x, u_{xx}, v, v_{x}, v_{xx}$).
\begin{equation}
\label{eqn:concentration1}
\begin{aligned}
 u_t(x,t) &= f^u(u, u_x, u_{xx},v, v_x, v_{xx}),\\
 v_t(x,t) &= f^v(u, u_x, u_{xx},v, v_x, v_{xx}).
\end{aligned}
\end{equation}
\begin{figure}[!htp]
  \centering
   \subfigure[~$u_t$ by GP]{
  \includegraphics[scale=0.17]{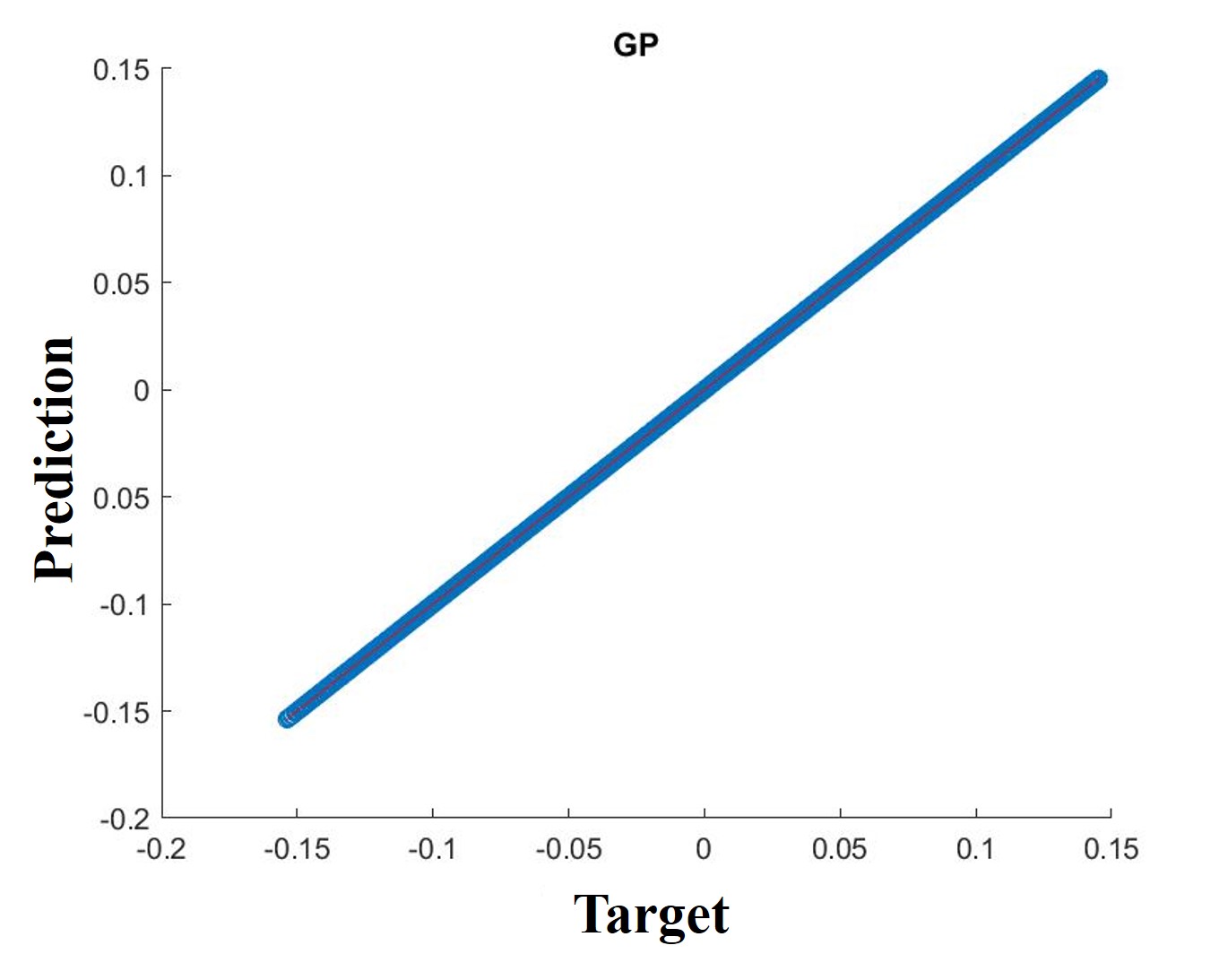}
  }
     \subfigure[~$v_t$ by GP]{
  \includegraphics[scale=0.17]{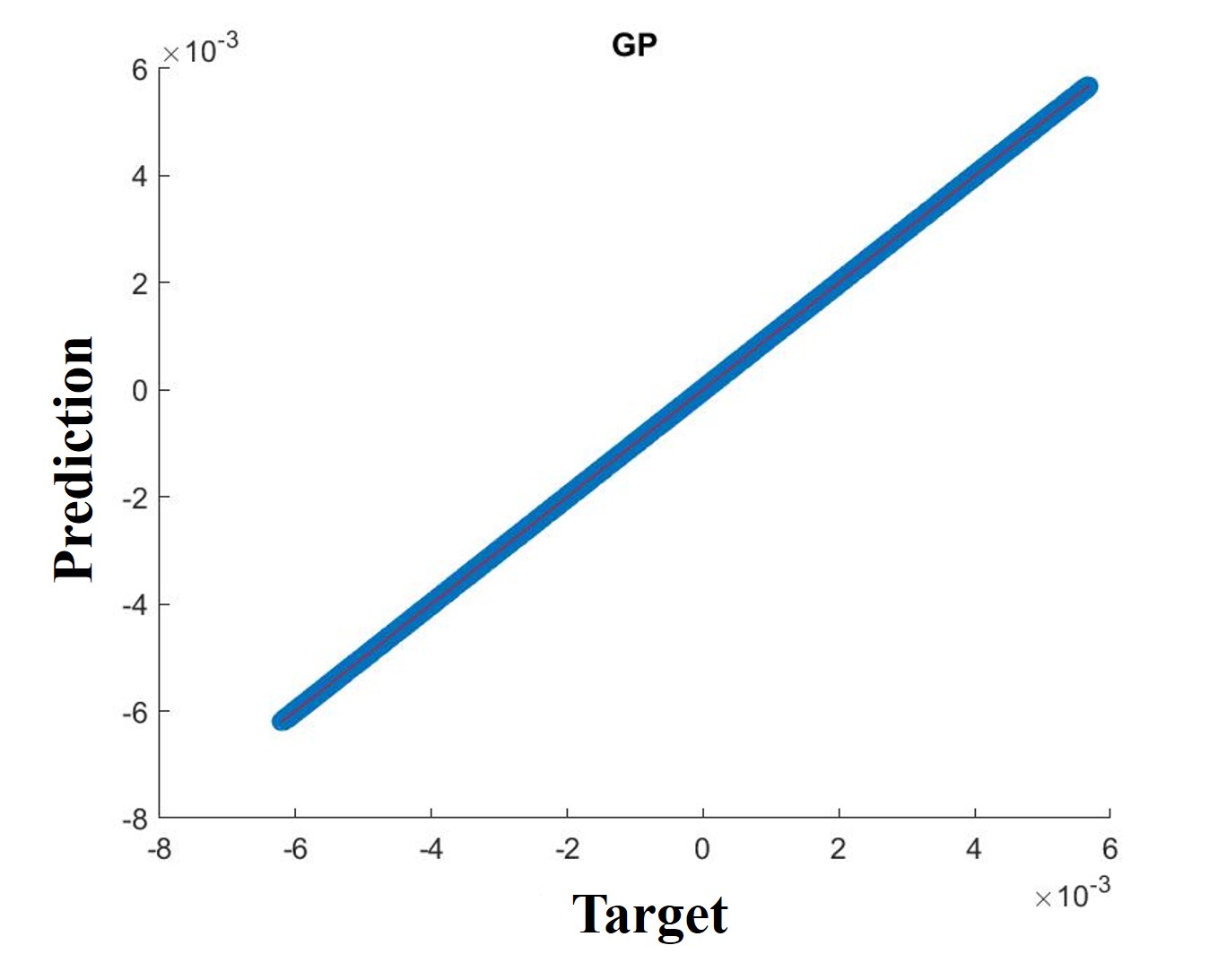}
  }
     \subfigure[~$u_t$ by NN]{
  \includegraphics[scale=0.17]{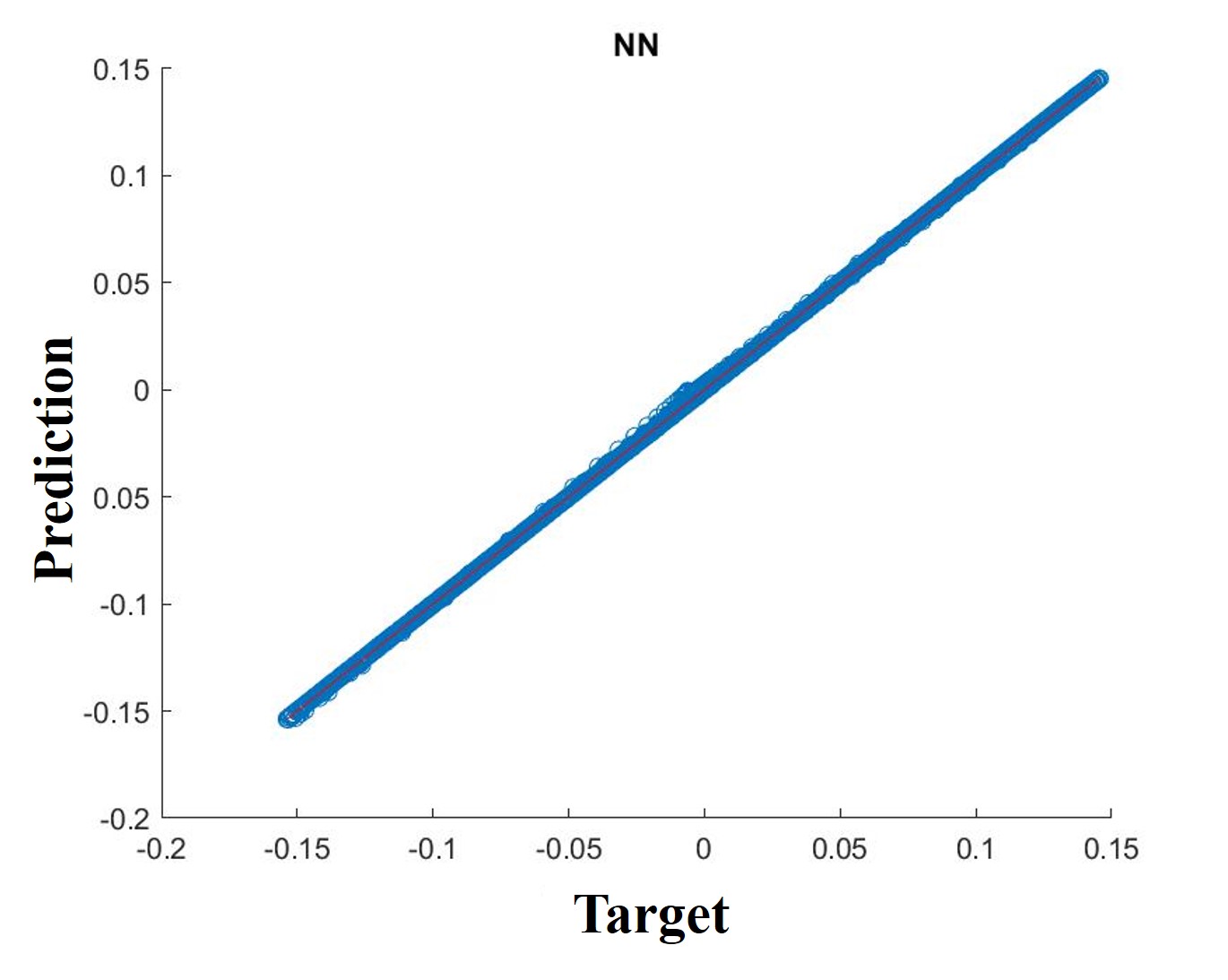}
  }
     \subfigure[~$v_t$ by NN]{
  \includegraphics[scale=0.17]{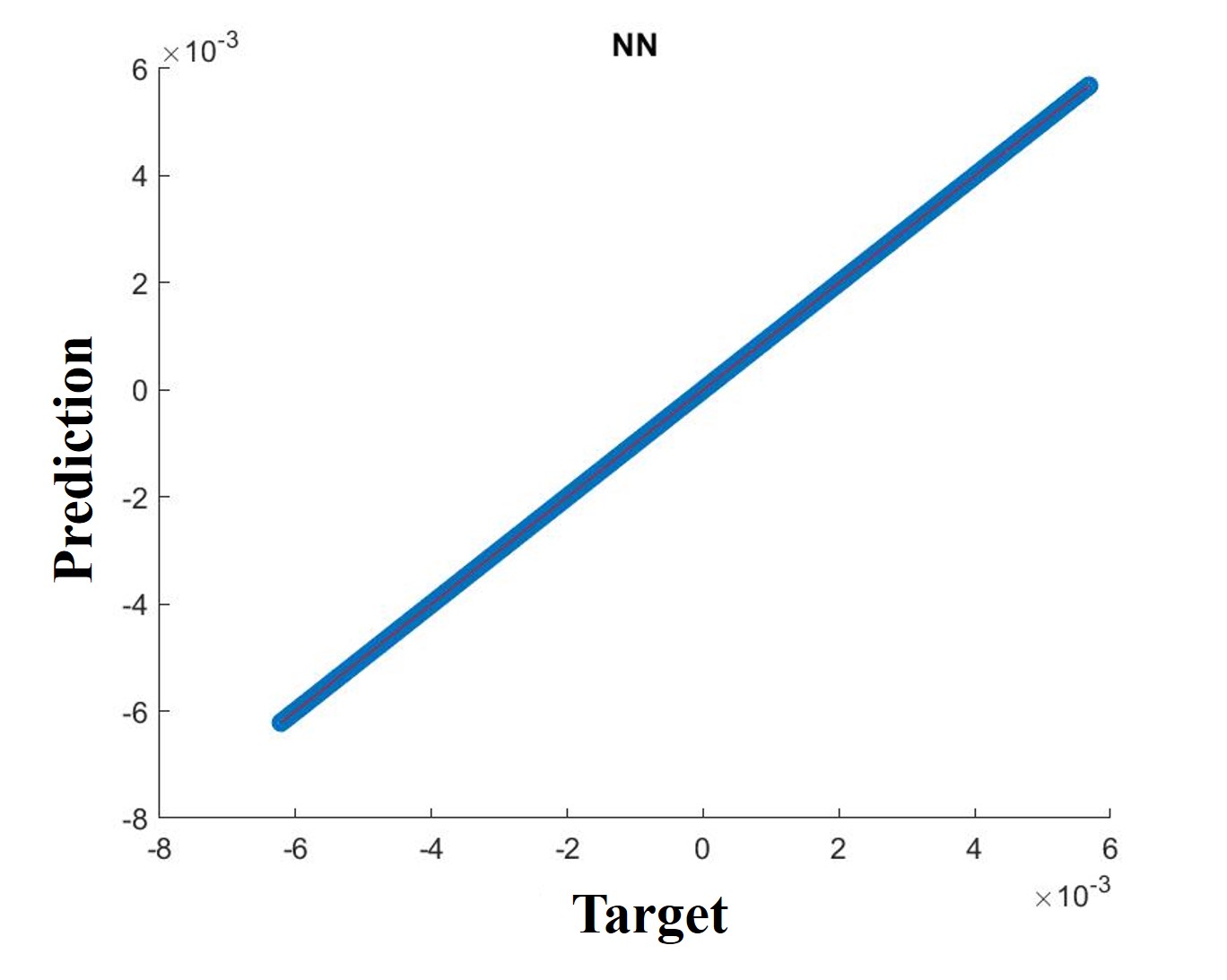}
  }
  \caption{No feature selection: $u_t=f^u(u,u_{x},u_{xx},v,v_x,v_{xx})$ and $v_t=f^v(u,u_{x},u_{xx},v,v_x,v_{xx})$.
  Regression results of the two methods for time derivatives: Gaussian processes (GP) and neural networks (NN). 
  }
  \label{fig:derivative_nf}
\end{figure}

\begin{figure}[!htp]
  \centering
    \subfigure[~Gaussian processes]{
  \includegraphics[scale=0.125]{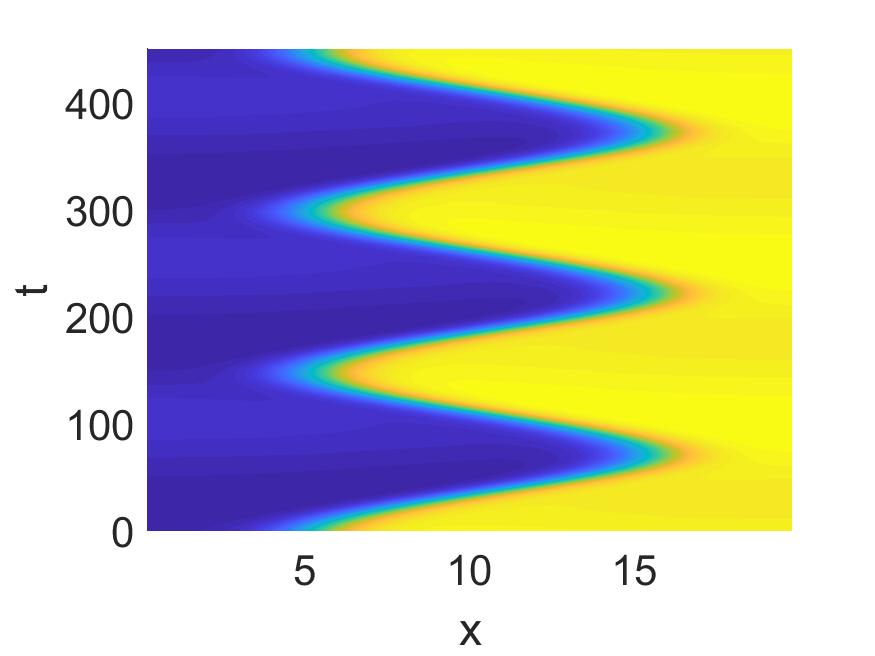}
  }
    \subfigure[~Neural networks]{
  \includegraphics[scale=0.125]{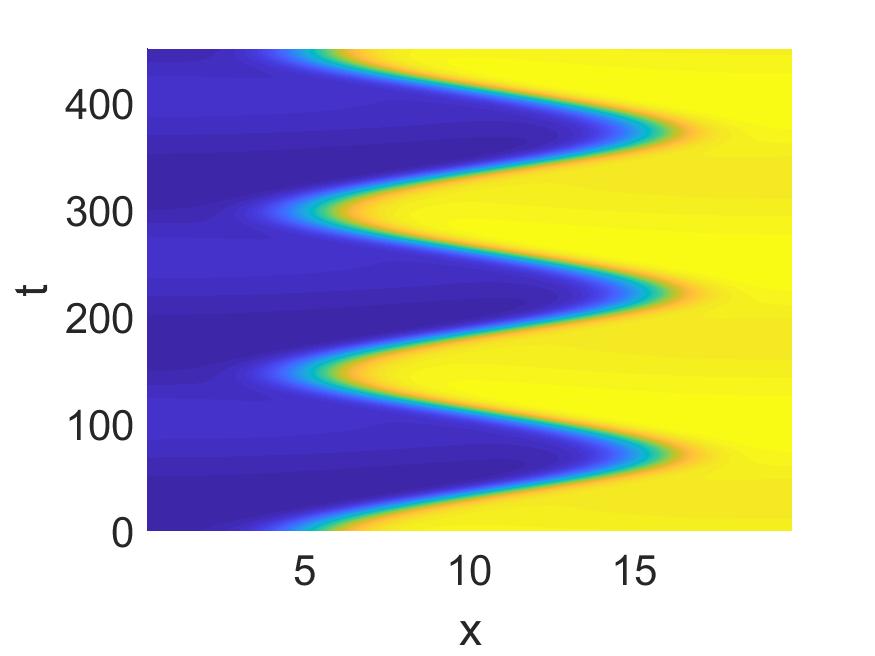}
  }  
     \subfigure[~Gaussian processes]{
  \includegraphics[scale=0.125]{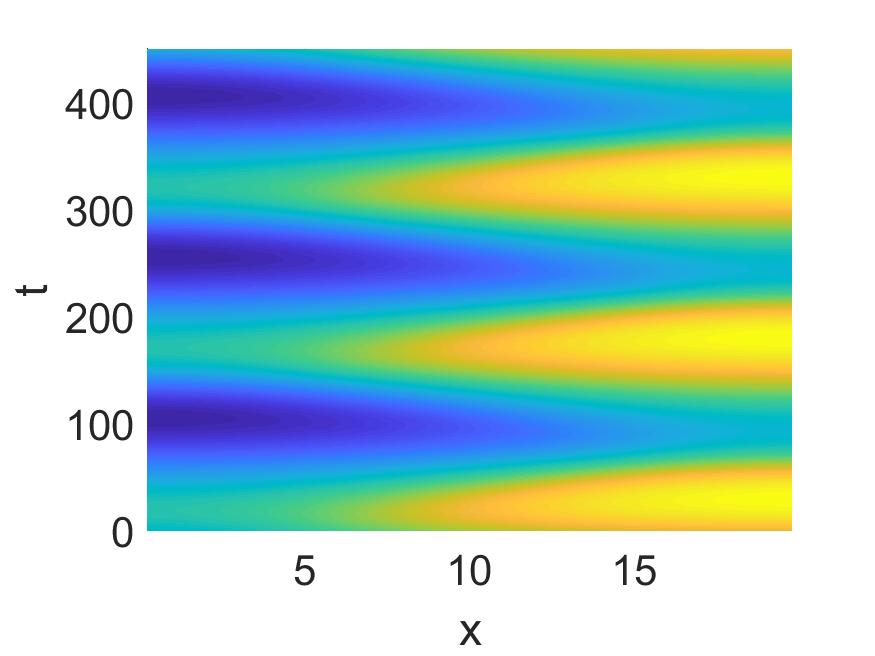}
  } 
      \subfigure[~Neural networks]{
  \includegraphics[scale=0.125]{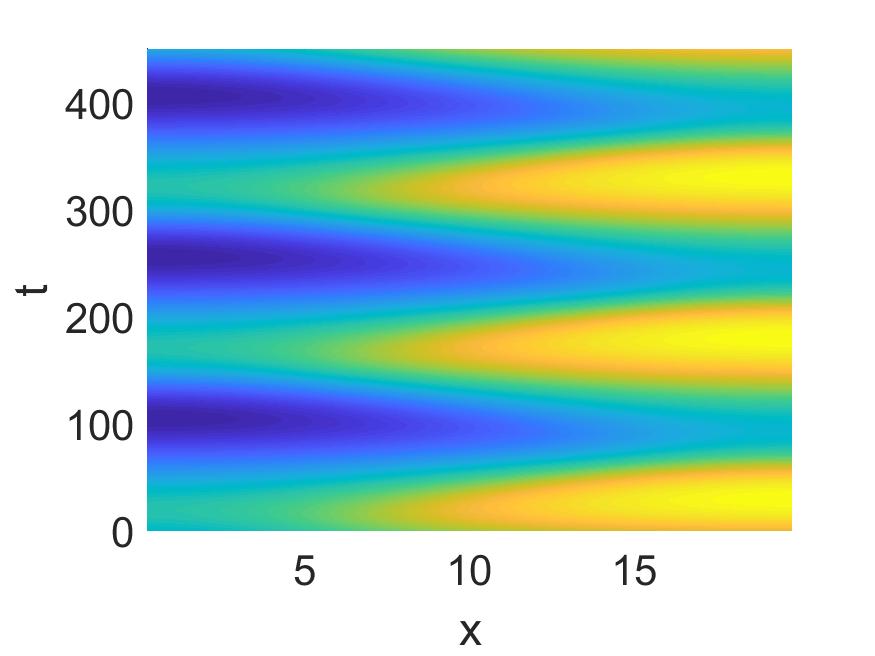}
  }
    \subfigure[~Absolute difference for $u$ (GP)]{
  \includegraphics[scale=0.125]{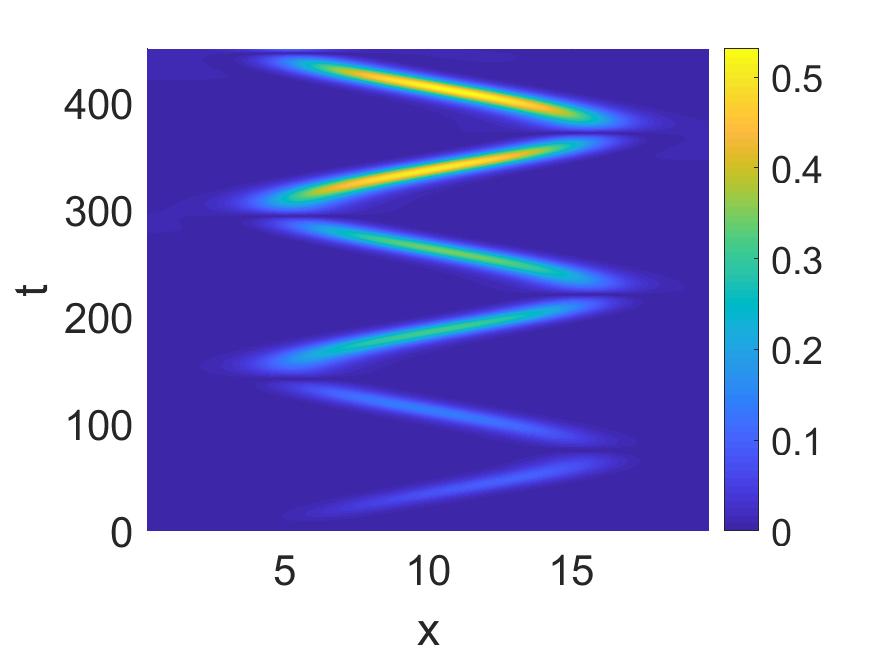}
  } 
    \subfigure[~Absolute difference for $u$ (NN)]{
  \includegraphics[scale=0.125]{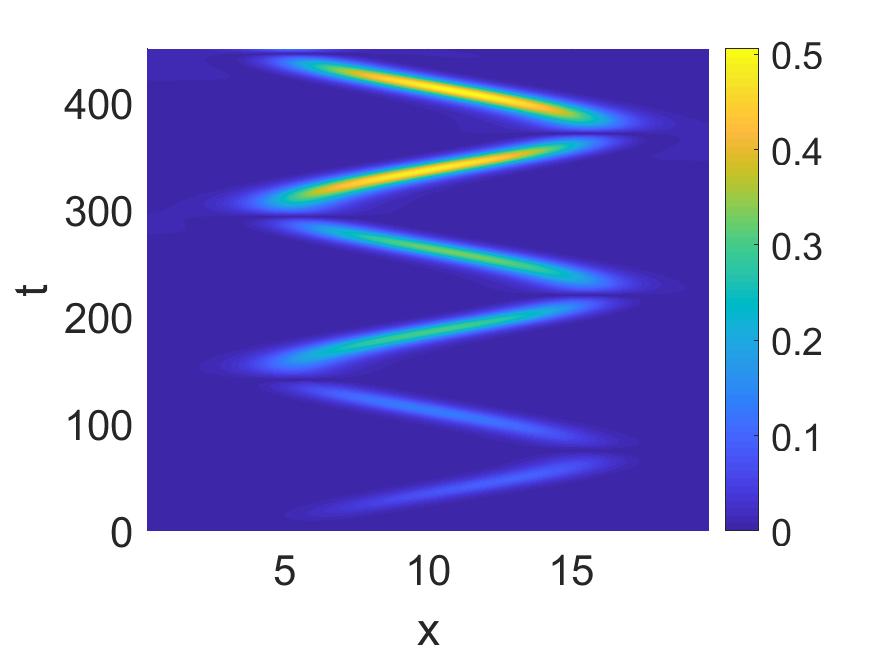}
  }
    \subfigure[~Absolute difference for $v$ (GP)]{
  \includegraphics[scale=0.125]{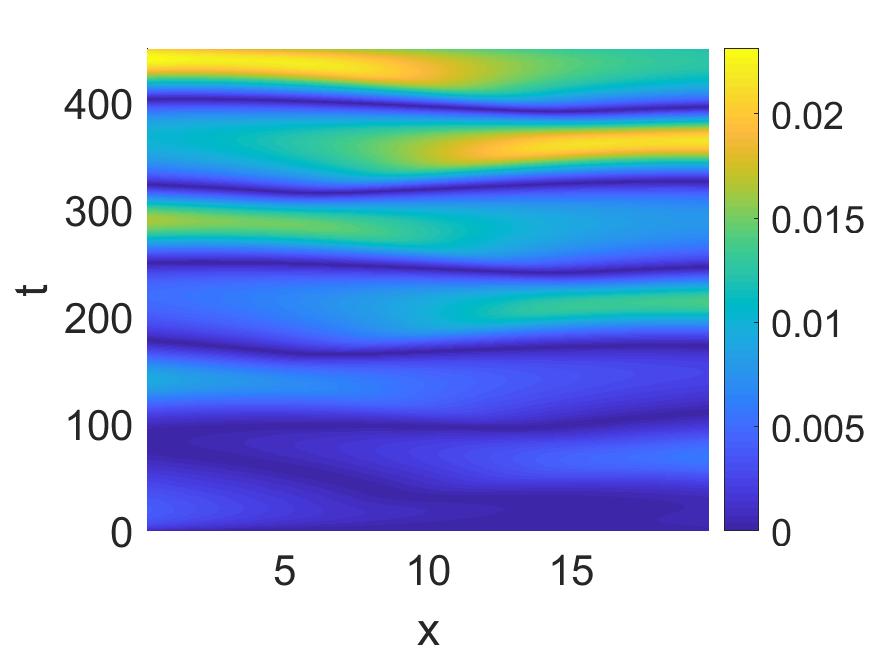}
  } 
    \subfigure[~Absolute difference for $v$ (NN)]{
  \includegraphics[scale=0.125]{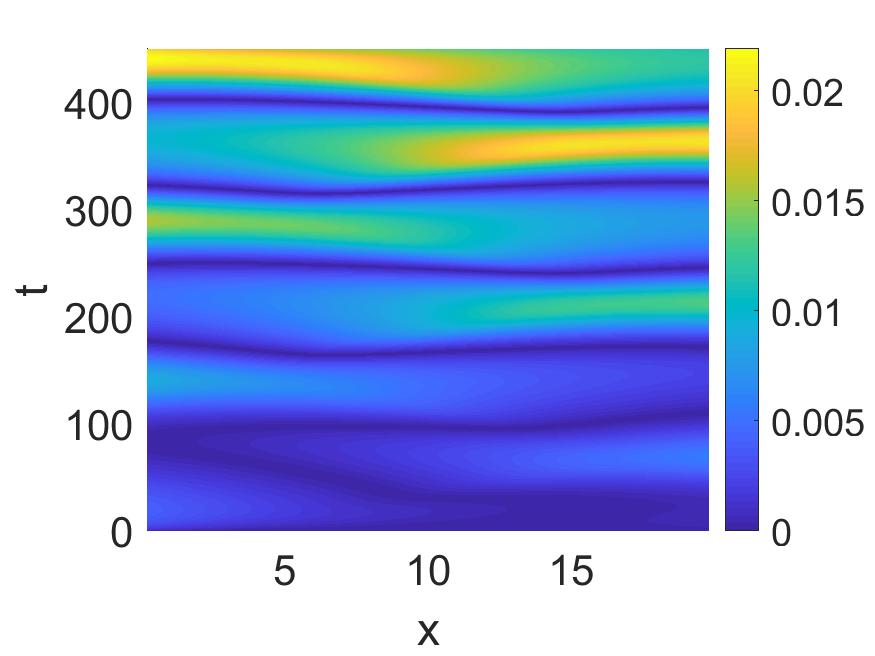}
  }  
  \caption{No feature selection: $u_t=f^u(u,u_{x},u_{xx},v,v_x,v_{xx})$ and $v_t=f^v(u,u_{x},u_{xx},v,v_x,v_{xx})$. (a)-(d): Simulation results of the two methods for $u$ and $v$. (e)-(h): The normalized absolute differences from the ``ground truth" LB simulations for $u$ and $v$.}
  \label{fig:nf}
\end{figure}

Specifically, for the neural networks approach, we build two different networks, one for the prediction of the activator and one for the inhibitor.
For both the activator and the inhibitor, we set use hidden layers consisting of 6 and 6 neurons using a hyperbolic tangent sigmoid activation function; as mentioned above, we use  Levenberg-Marquardt optimization with a Bayesian regularization (see section~\ref{sec:ANN}).
Both networks use the mean-squared-error as their loss function.
For Gaussian processes, we employ a radial basis kernel function with ARD (see equation~\eqref{eqn:kernel}).
Regression results obtained by each the two methods for the time derivatives in the training data set are shown in figure~\ref{fig:derivative_nf}.
Both methods provide good approximations of the target time derivatives $u_t$ and $v_t$.
Given the test coarse initial condition (black curves in figure~\ref{fig:ini}), simulation results {\em with the learned PDE} from $t=0$ to $t=450$ with $\Delta t=0.001$ and their normalized absolute differences from the ``ground truth" LB simulations are shown in figures~\ref{fig:nf}.
The order of magnitude of these absolute differences for both models is the same as those between the LB FHN and the explicitly known FHN PDE (see figures~\ref{fig:origLBM}(e) and (f)). 
%
\subsection{Learning with feature selection}
\label{sec:fs}
\begin{figure}[!htp]
  \centering
   \subfigure[~$u_t$ by GP]{
  \includegraphics[scale=0.17]{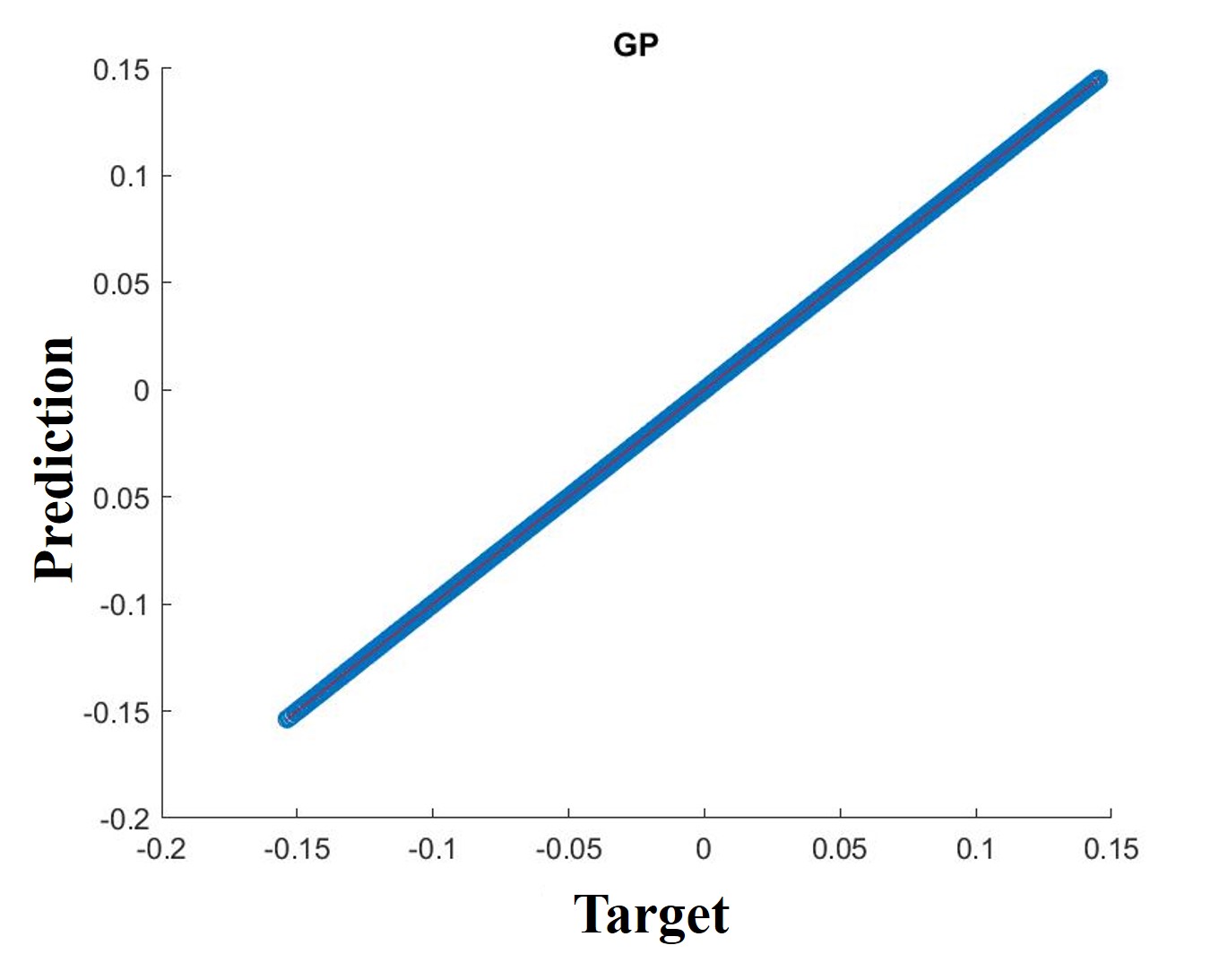}
  }
     \subfigure[~$v_t$ by GP]{
  \includegraphics[scale=0.17]{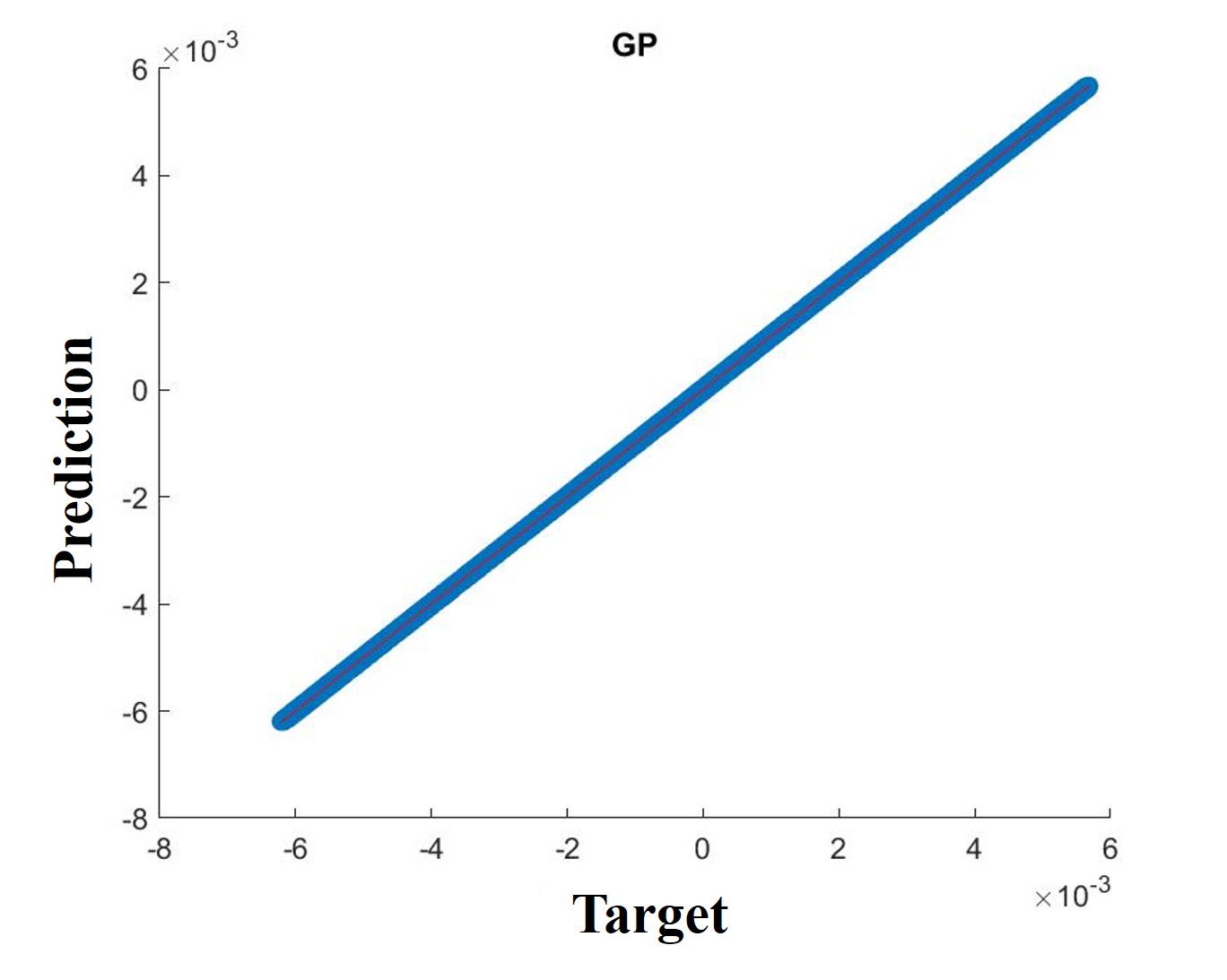}
  }
     \subfigure[~$u_t$ by NN]{
  \includegraphics[scale=0.17]{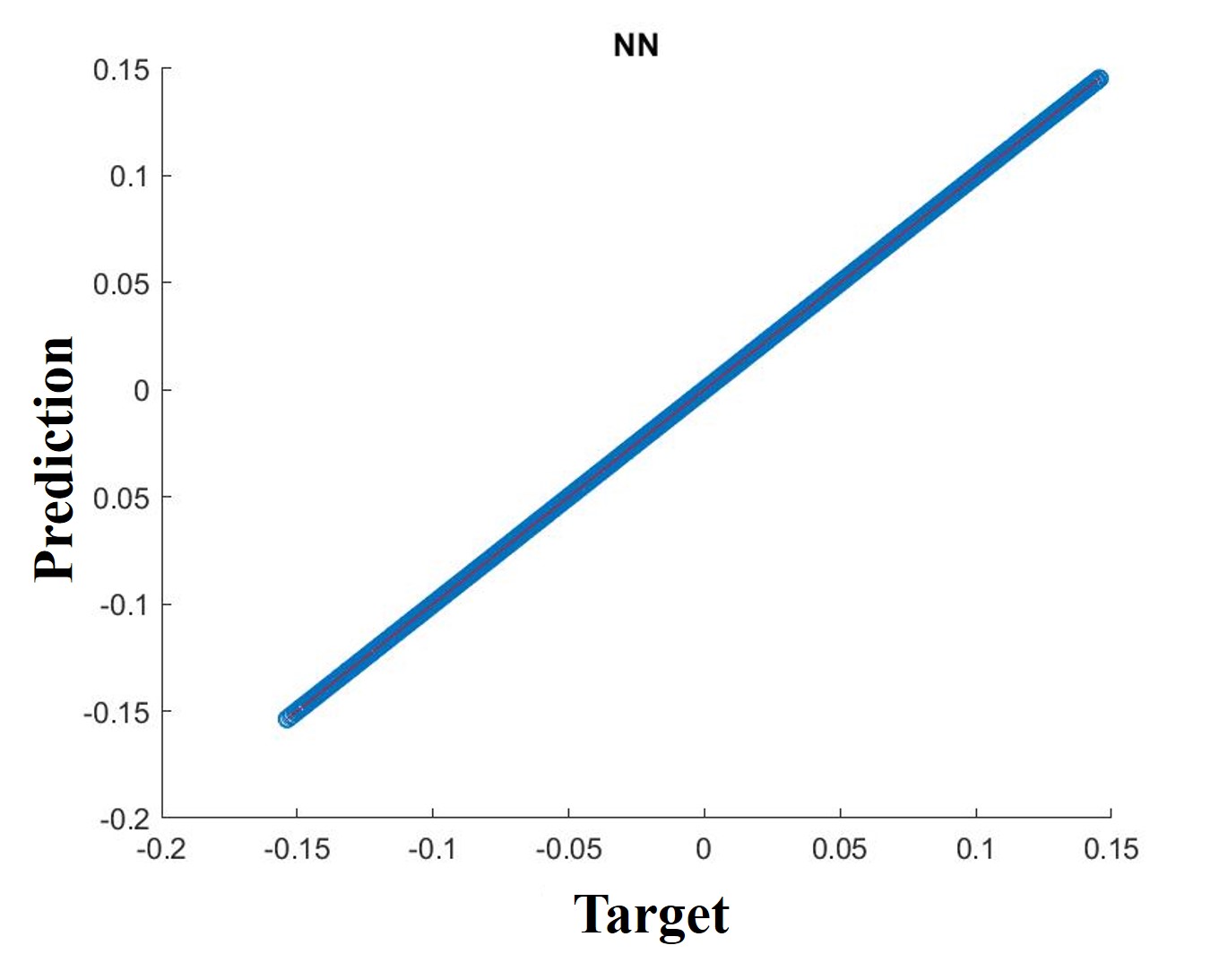}
  }
     \subfigure[~$v_t$ by NN]{
  \includegraphics[scale=0.17]{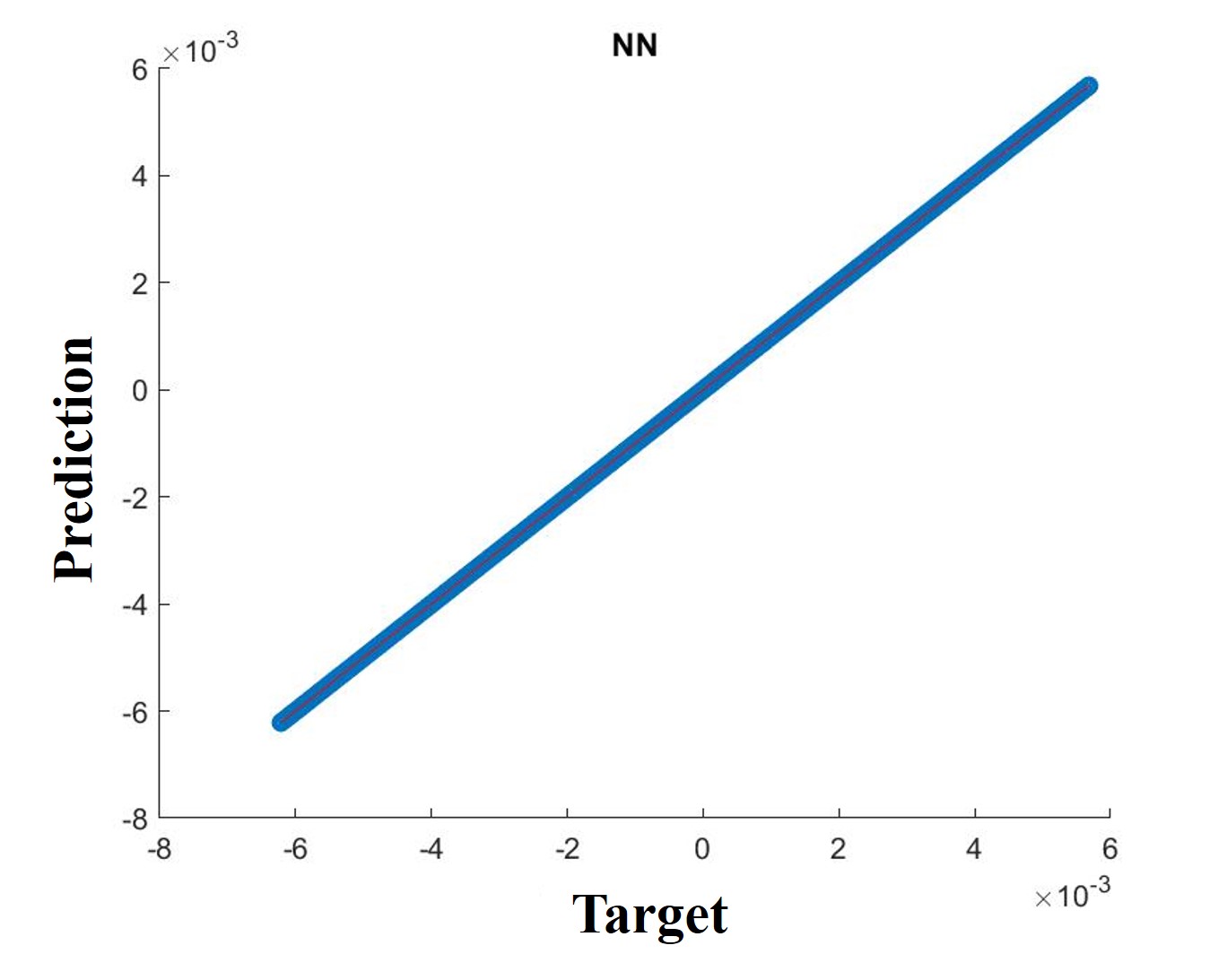}
  }
  \caption{Feature selection 1: $u_t=f_1^u(u,u_{xx},v)$ and $v_t=f_1^v(u,v,v_{xx})$.
  These selected variables are the same as those that appear in the right-hand-side of the explicitly known FHN PDE.
  Regression results of the two methods for time derivatives: Gaussian processes (GP) and neural networks (NN).
  }
  \label{fig:derivative_f1}
\end{figure}

\begin{figure}[!htp]
  \centering
    \subfigure[~Gaussian processes]{
  \includegraphics[scale=0.125]{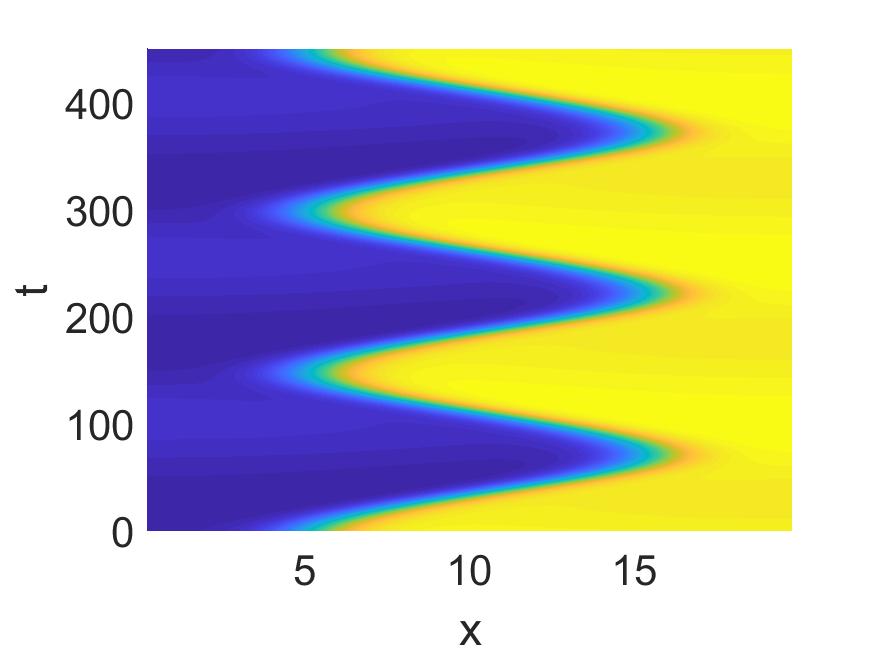}
  }
    \subfigure[~Neural networks]{
  \includegraphics[scale=0.125]{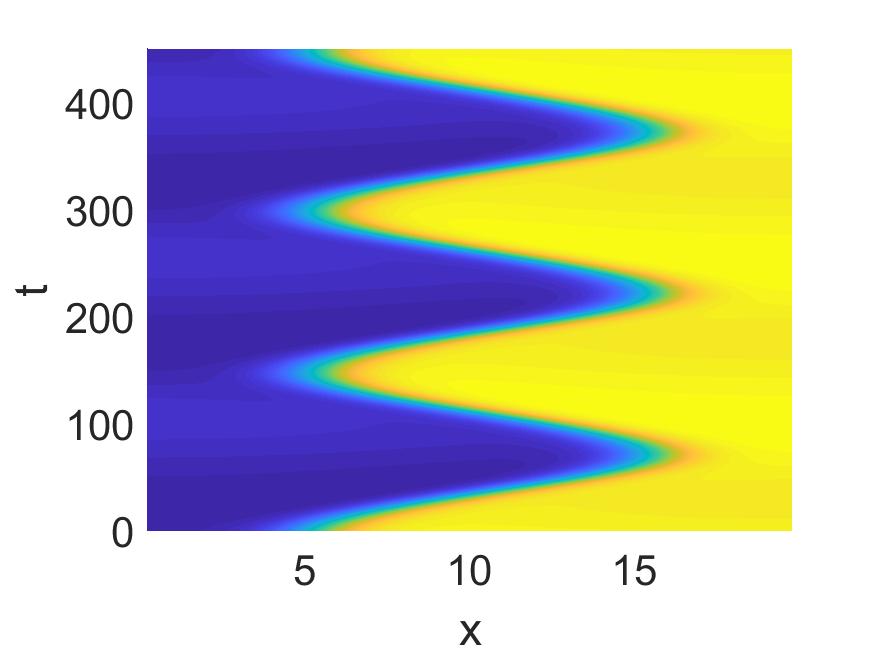}
  }  
     \subfigure[~Gaussian processes]{
  \includegraphics[scale=0.125]{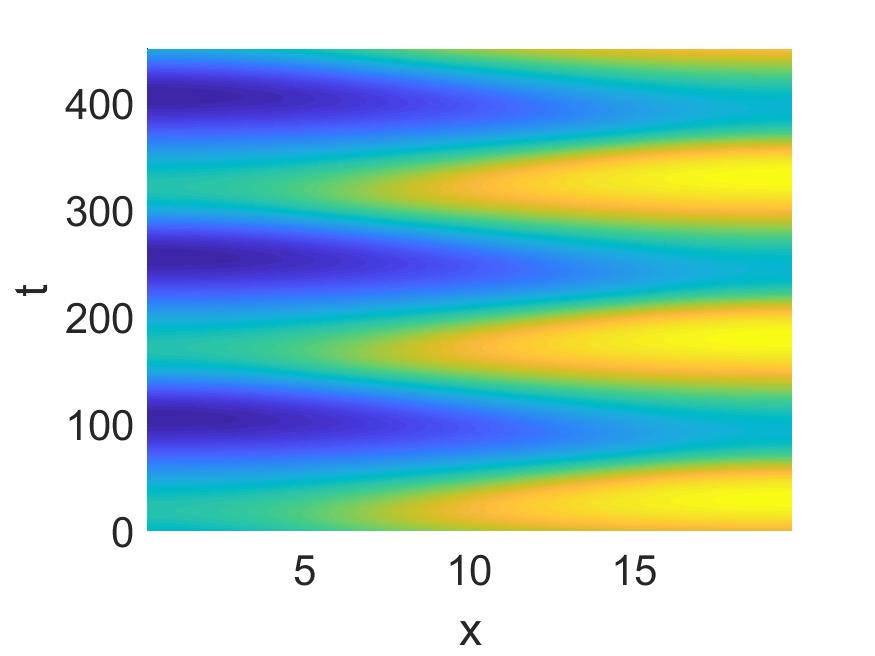}
  } 
      \subfigure[~Neural networks]{
  \includegraphics[scale=0.125]{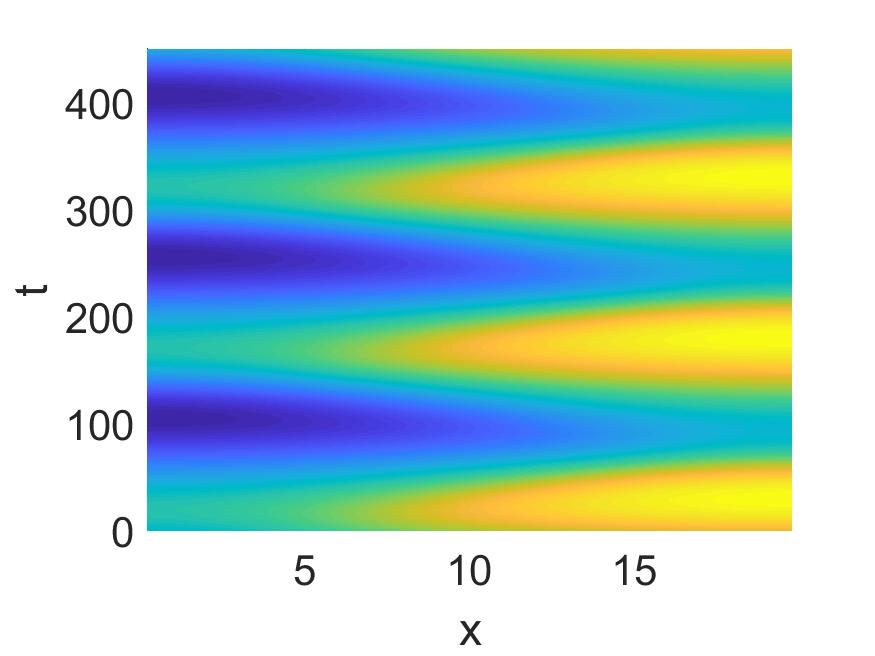}
  }
    \subfigure[~Absolute difference for $u$ (GP)]{
  \includegraphics[scale=0.125]{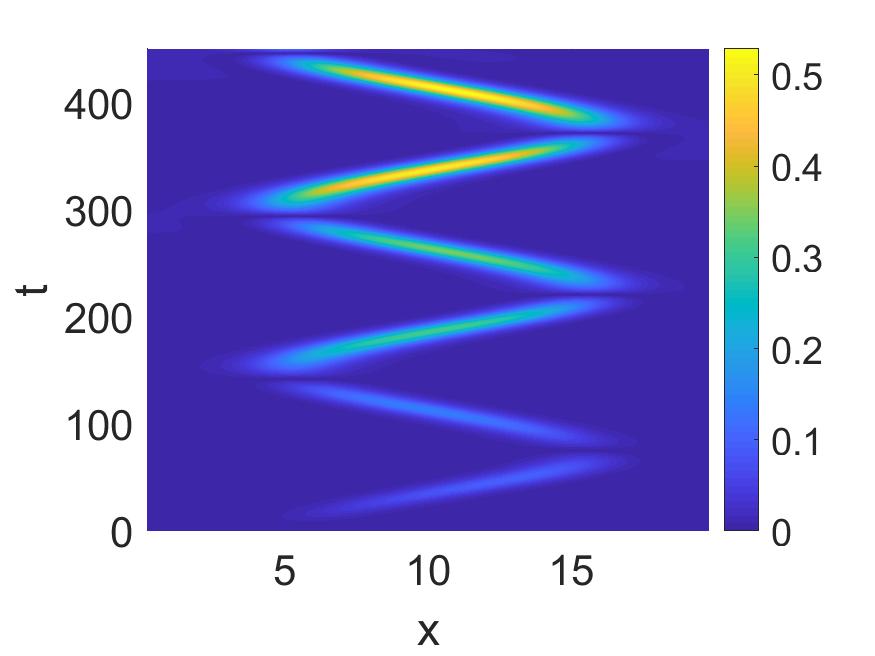}
  } 
    \subfigure[~Absolute difference for $u$ (NN)]{
  \includegraphics[scale=0.125]{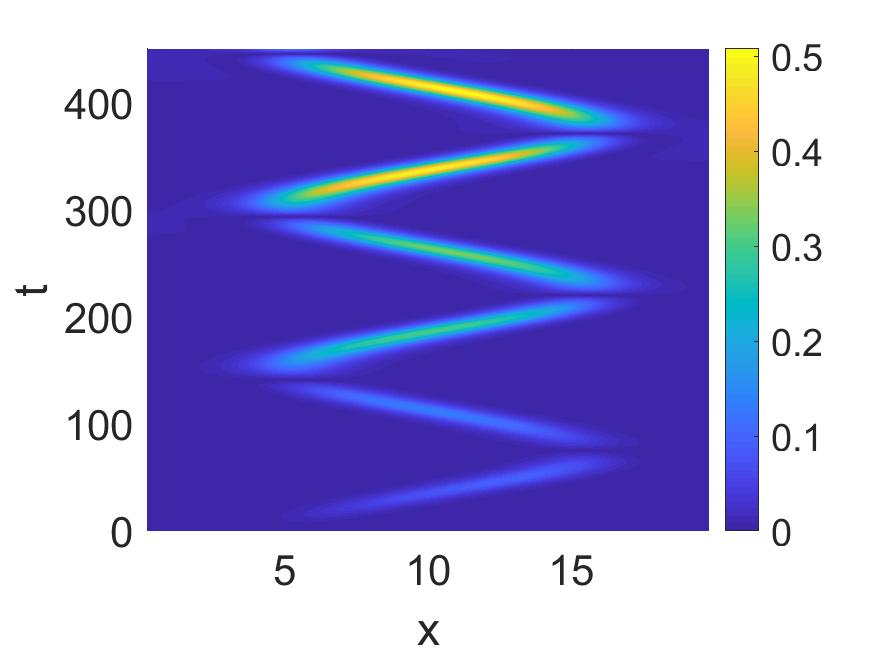}
  }
    \subfigure[~Absolute difference for $v$ (GP)]{
  \includegraphics[scale=0.125]{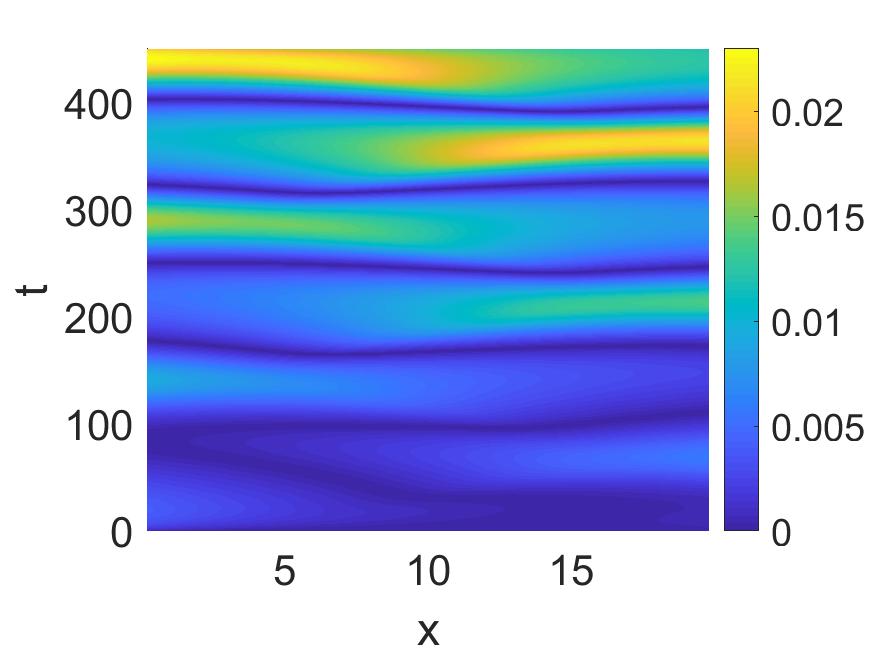}
  } 
    \subfigure[~Absolute difference for $v$ (NN)]{
  \includegraphics[scale=0.125]{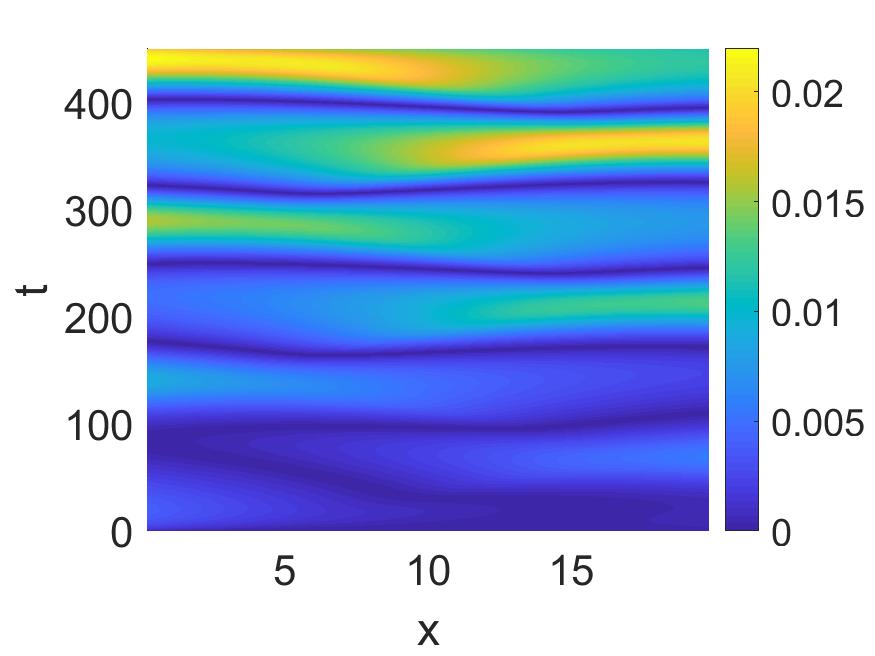}
  }  
  \caption{Feature selection 1: $u_t=f^u(u,u_{xx},v)$ and $v_t=f^v(u,v,v_{xx})$.
  (a)-(d): Simulation results of the two methods for $u$ and $v$. (e)-(h): The normalized absolute differences from the ``ground truth" LB simulations for $u$ and $v$.
  }
  \label{fig:f1}
\end{figure}

Now, we consider the possibility of feature selection, in an attempt to learn the RHS of coarse-scale PDEs with a minimal number of input domain variables (spatial derivatives). 
First, we apply the sensitivity analysis via ARD in the case of Gaussian process approximation.
The optimal ARD weights ($\theta^*$) for $u_t$ and $v_t$ are tabulated in table~\ref{tab:ard}.
$u_t$ has three relatively small weights for ($u, u_{xx}, v$) and $v_t$ has also three relatively small weights for ($u, v, v_{xx}$).
It is interesting to observe that the selected features via ARD are the same as those in the explicitly known FHN PDE (see equation~\eqref{eqn:fhn}). 
This shows that ARD can effectively guide in selecting the appropriate dimensionality of the input data domain, resulting here in the same spatial derivative choices as in the explicitly known FHN PDE.
\begin{table}[!htp]
\caption{\label{tab:ard} Optimal ARD weights ($\theta^*$ for $u_t$ and $v_t$ in equation~\eqref{eqn:opt}). As mentioned in section~\ref{sec:FS}, features which have relatively small ARD weights can be regarded as dominant features for the target functions $u_t$ and $v_t$.}
\begin{ruledtabular}
\begin{tabular}{ccccccc}
            & $u$ & $u_x$ & $u_{xx}$ & $v$ & $v_x$ & $v_{xx}$ \\
\hline
    $u_t$   & 5.28E+00  & 4.23E+06 & 9.13E+02 & 2.13E+03 & 5.32E+08 & 4.78E+07 \\
    $v_t$   & 1.33E+02  & 6.69E+06 & 1.94E+06 & 5.09E+02 & 4.20E+06 & 1.75E+02\\
\end{tabular}
\end{ruledtabular}
\end{table}
Now, we use the reduced input data domain ($u, u_{xx}, v$) for $u_t$ and ($u, v, v_{xx}$) for $v_t$ to recover the RHS of the coarse-scale PDEs as
\begin{equation}
\label{eqn:f1}
\begin{aligned}
 u_t(x,t) &= f_1^u(u, u_{xx},v),\\
 v_t(x,t) &= f_1^v(u,v, v_{xx}).
\end{aligned}
\end{equation}
Regression results of our two methods for the time derivatives are shown in figure~\ref{fig:derivative_f1}.
Results of long time simulation of the learned PDEs by each method, from $t=0$ to $t=450$, as well as normalized absolute differences from the simulation of the ``ground truth" LB are shown in figure~\ref{fig:f1}.
%

The two machine learning methods operating with a reduced input data domain can still provide good approximations of the time derivatives and of the resulting dynamics.
The order of magnitude of these absolute differences is effectively the same as the difference of the FHN LB from the explicitly known FHN PDE.
It is, therefore, clear that our framework effectively recovers the coarse-scale PDE from fine scale observation data; the difference is that the right hand-side of the PDE is now given in terms of the ANN right-hand-side, or in terms of the observed data and the GP kernel/hyperparameters, rather than the simple algebraic formula of equation \eqref{eqn:fhn}.
\begin{table}[!htp]
\caption{The best candidates and the corresponding regression loss (L) for $u_t$ and $v_t$ with respect to the number of Diffusion map coordinates}
\label{tab:dimu}
\begin{ruledtabular}
\begin{tabular}{cllcc}
    & \multicolumn{2}{l}{Optimal intrinsic coordinates}  &\multicolumn{2}{l}{Regression Loss (L)} \\
    &  $u_t$ & $v_t$ & $u_t$ & $v_t$ \\ \hline
    1d  &  ($\phi^u_5$) & ($\phi^v_2$) & 4.60E-04& 7.69E-06 \\
    2d  & ($\phi^u_1,\phi^u_5$) & ($\phi^v_1,\phi^v_2$) &1.40E-06 &1.50E-06\\
    3d  & ($\phi^u_1,\phi^u_4,\phi^u_5$)   &($\phi^v_1,\phi^v_2,\phi^v_3$) &2.18E-08 &4.74E-08\\
    4d  & ($\phi^u_1,\phi^u_3,\phi^u_4,\phi^u_5$) & ($\phi^v_1,\phi^v_2,\phi^v_3,\phi^v_4$) &1.64E-08 &5.71E-09\\
\end{tabular}
\end{ruledtabular}
\end{table}
\begin{table}[!htp]
\caption{The best candidates and corresponding total loss for $u_t=(\phi^u_1,\phi^u_4,\phi^u_5)$ and $v_t=(\phi^v_1,\phi^v_2,\phi^v_3)$ with respect to the number of features.}
 \label{tab:f1u}
\begin{ruledtabular}
\begin{tabular}{clclc}
    & \multicolumn{2}{c}{$u_t=(\phi^u_1,\phi^u_4,\phi^u_5)$}  &\multicolumn{2}{c}{$v_t=(\phi^v_1,\phi^v_2,\phi^v_3)$} \\
    &  Features & Total Loss ($L_T$) & Features & Total Loss ($L_T$) \\ \hline
    1d  &  ($u$) & 6.51E-05 &  ($u$) & 7.93E-05 \\
    2d  & ($u,v$) & 1.65E-08 &($u,v$) &1.49E-05 \\
    3d  & ($u, u_{xx}, v$)   &6.52E-09 &($u, v, v_{xx}$) &3.32E-07\\
        & ($u, u_{x}, v$)   &7.39E-09 &($u, u_{x}, v_{xx}$) &6.21E-07\\
    4d  & ($u, u_{x},u_{xx}, v$) & 2.68E-09 &($u, v, v_{x}, v_{xx}$)& 4.47E-09\\
\end{tabular}
\end{ruledtabular}
\end{table}
\begin{figure}[!htp]
  \centering
   \subfigure[~$u_t = f_{ud}(\phi^u_1, \phi^u_4, \phi^u_5)$]{
  \includegraphics[scale=0.2]{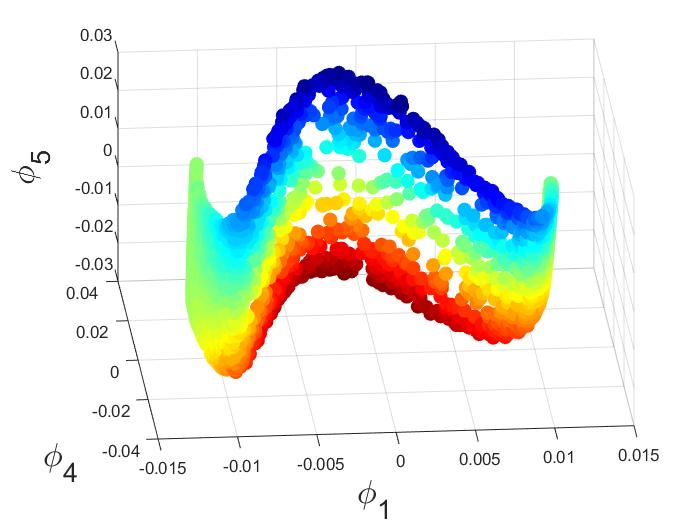}
  }
     \subfigure[~$v_t = f_{vd}(\phi^v_1, \phi^v_2, \phi^v_3)$]{
  \includegraphics[scale=0.2]{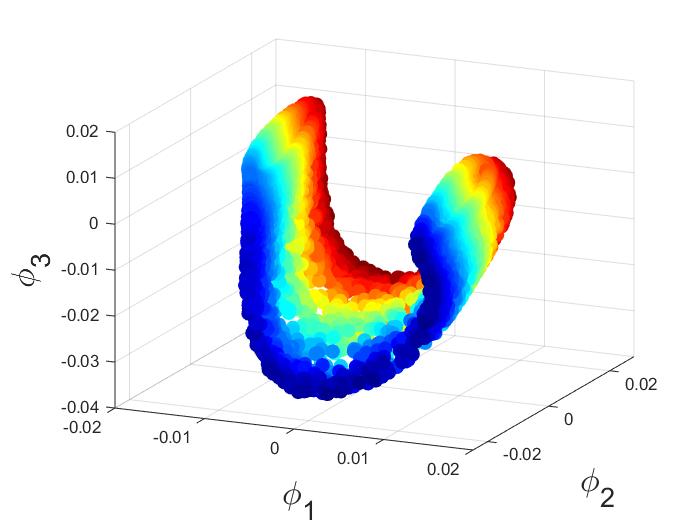}
  }

  \caption{Three leading Diffusion map coordinates: Colors represent $u_t$ in (a) and $v_t$ in (b).}
  \label{fig:dmap}
\end{figure}

Next, we consider an alternative approach for feature selection, via our manifold learning technique, Diffusion Maps.
The best candidate set among different combinations of intrinsic coordinates (varying the number of leading intrinsic dimensions and recording the corresponding Gaussian Process regression loss) are shown in table~\ref{tab:dimu}.
Since this three-dimensional intrinsic embedding space exhibits a (tiny) regression loss of order $10^{-8}$, we choose an input domain for $u_t$ consisting of ($\phi^u_1,\phi^u_4,\phi^u_5$) as shown in figure~\ref{fig:dmap}(a).
For $v_t$, by the same token, we choose the three-dimensional embedding space consisting of ($\phi^v_1,\phi^v_2,\phi^v_3$) as shown in figure~\ref{fig:dmap}(b).
Based on these identified intrinsic embedding spaces, we examined several subsets of input domain features (spatial derivatives) using the total loss of equation~\eqref{eqn:tloss}.
``Good" subsets of input features (those that result in small regression losses with minimal input dimension)
are presented in table~\ref{tab:f1u}.
Clearly, different choices of such input feature subsets can give comparable total losses; this suggests that we may construct different right-hand-sides of the unknown coarse-scale PDE that are comparably successful in representing the observed dynamics. 

The good candidates for $u_t$ and $v_t$ identified this way, consisting of three input features, are ($u, u_{xx}, v$) and ($u, v, v_{xx}$); they are  the same as those found from GP via ARD, and also the same as the ones in the explicitly known FHN PDE.
Interestingly, another possible alternative candidate set is also identified: ($u, u_{x}, v$) for $u_t$ and ($u, u_{x}, v_{xx}$) for $v_t$.
\begin{figure}[!htp]
  \centering
   \subfigure[~$u_t$ by GP]{
  \includegraphics[scale=0.17]{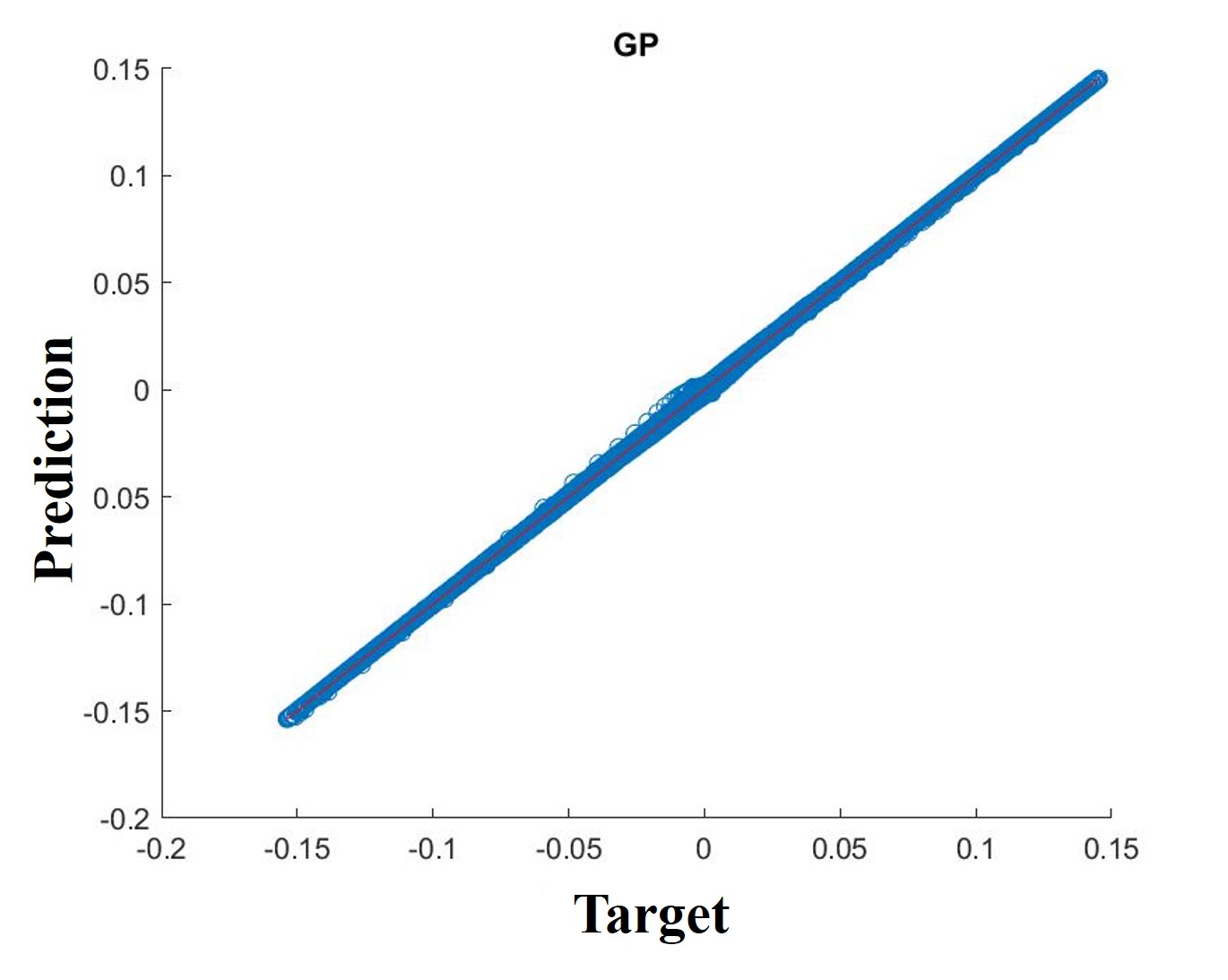}
  }
     \subfigure[~$v_t$ by GP]{
  \includegraphics[scale=0.17]{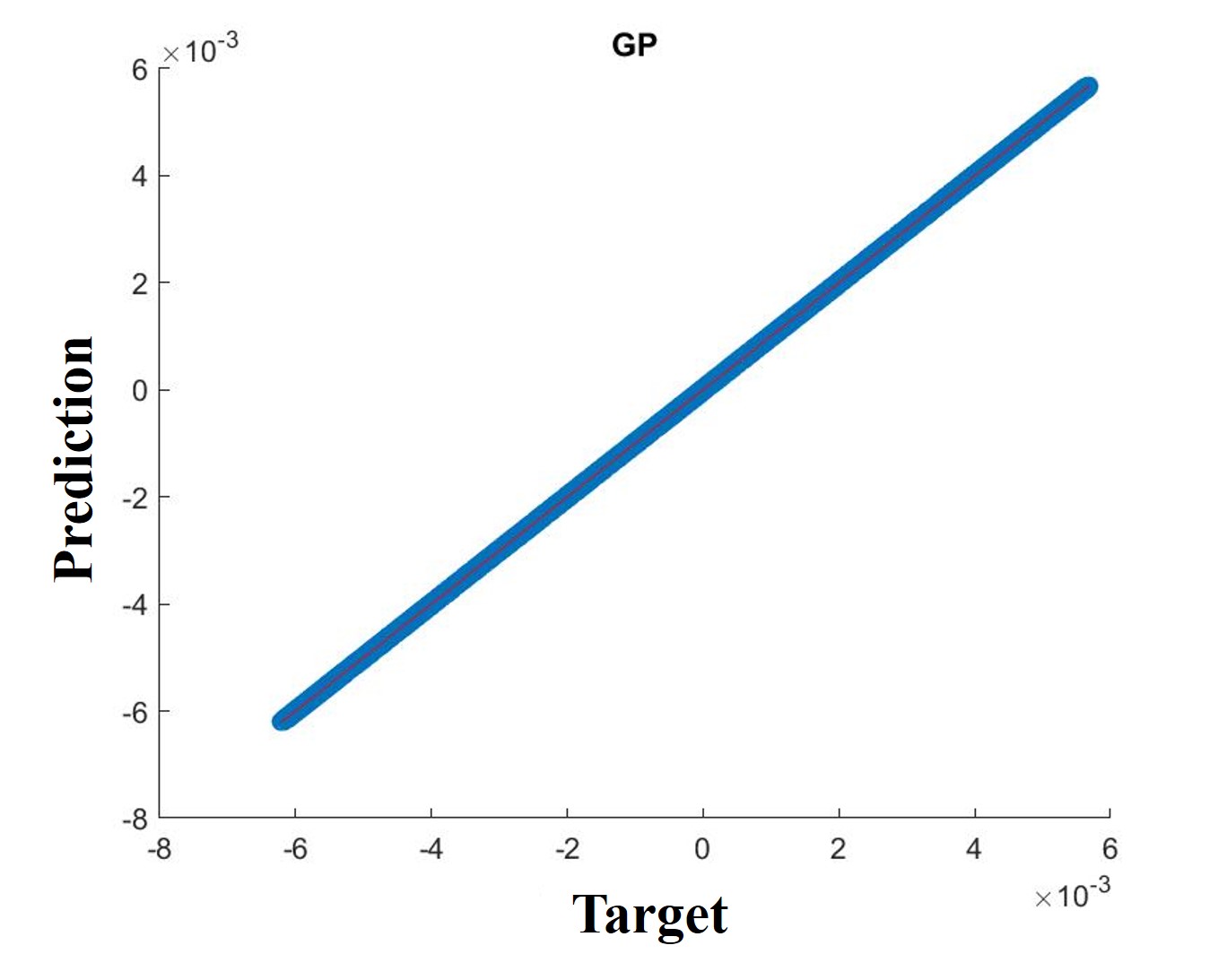}
  }
     \subfigure[~$u_t$ by NN]{
  \includegraphics[scale=0.17]{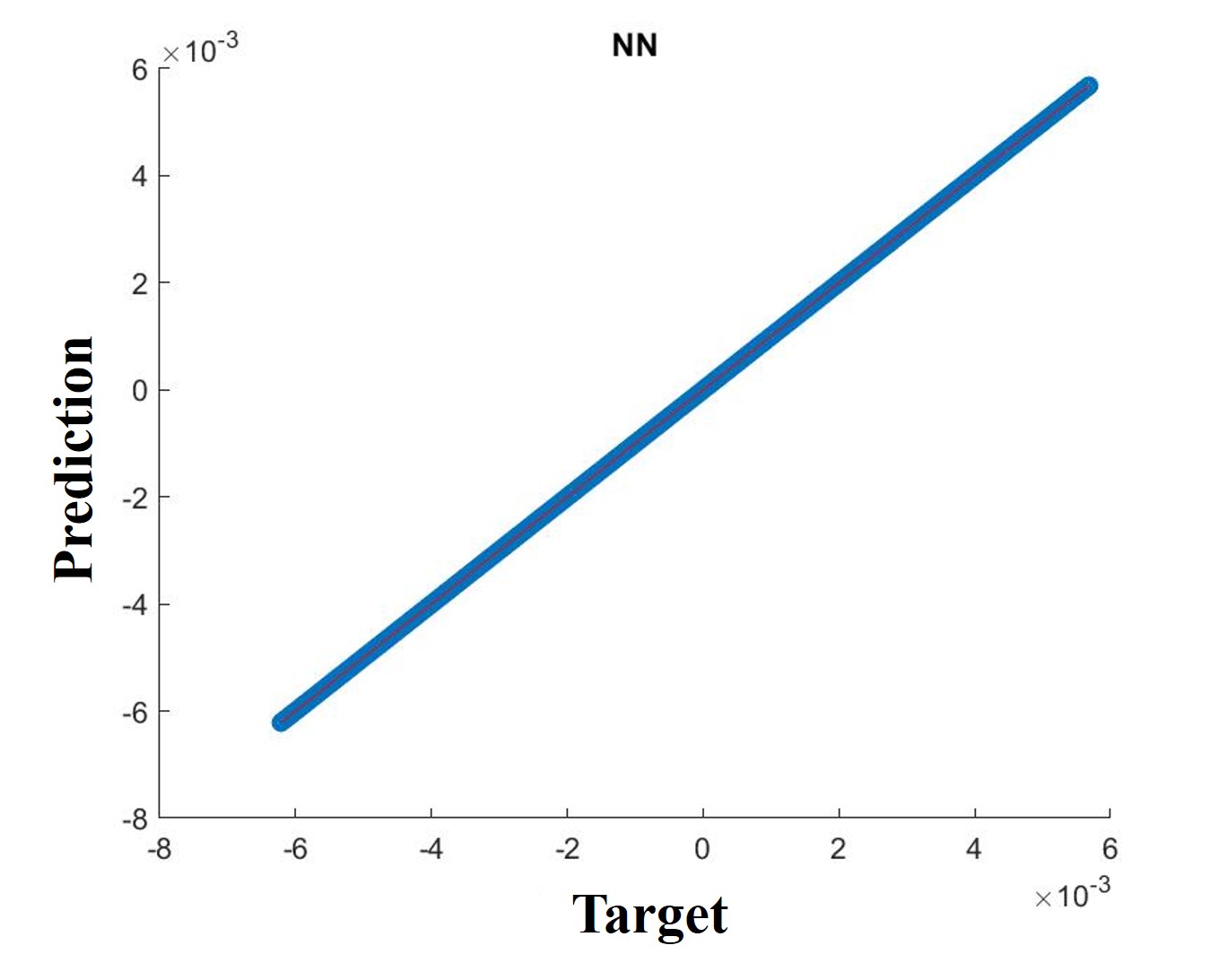}
  }
     \subfigure[~$v_t$ by NN]{
  \includegraphics[scale=0.17]{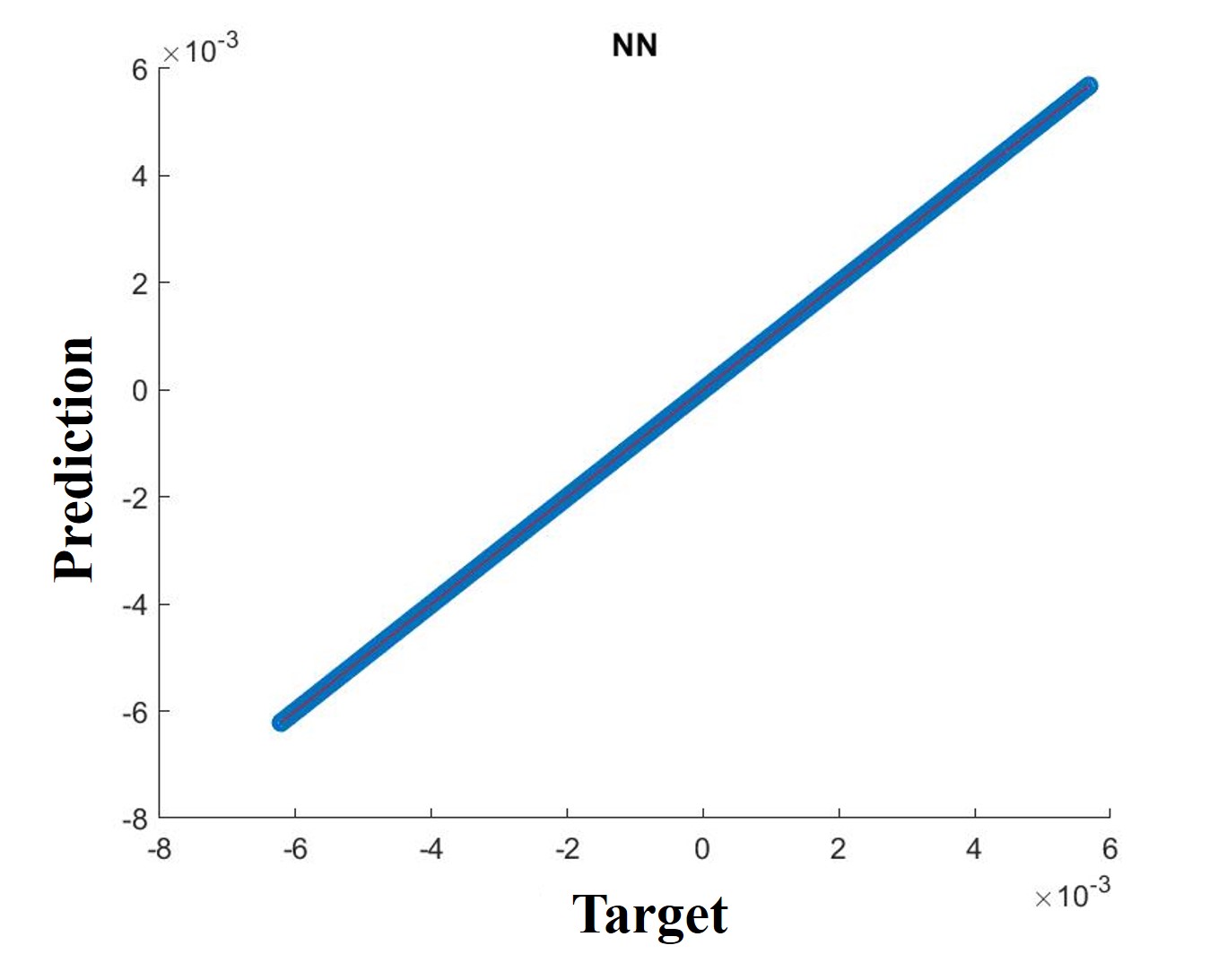}
  }
  \caption{Feature selection 2: $u_t=f_2^u(u,u_{x},v)$ and $v_t=f_2^v(u,v,v_{xx})$.
  Regression results of the two methods for time derivatives: Gaussian processes (GP) and neural networks (NN).
  }
  \label{fig:derivative_f2}
\end{figure}
\begin{figure}[!htp]
  \centering
    \subfigure[~Gaussian processes]{
  \includegraphics[scale=0.125]{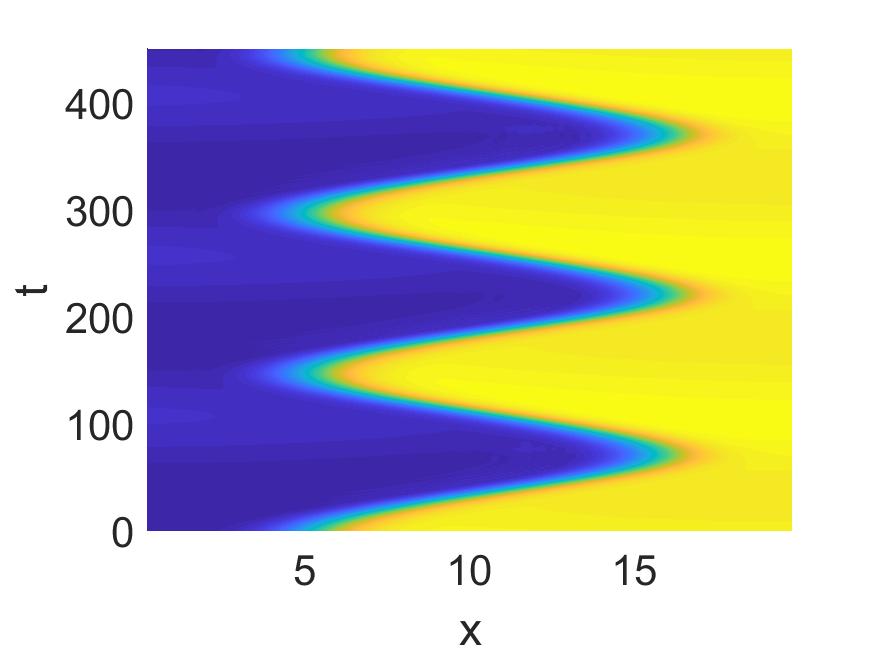}
  }
    \subfigure[~Neural networks]{
  \includegraphics[scale=0.125]{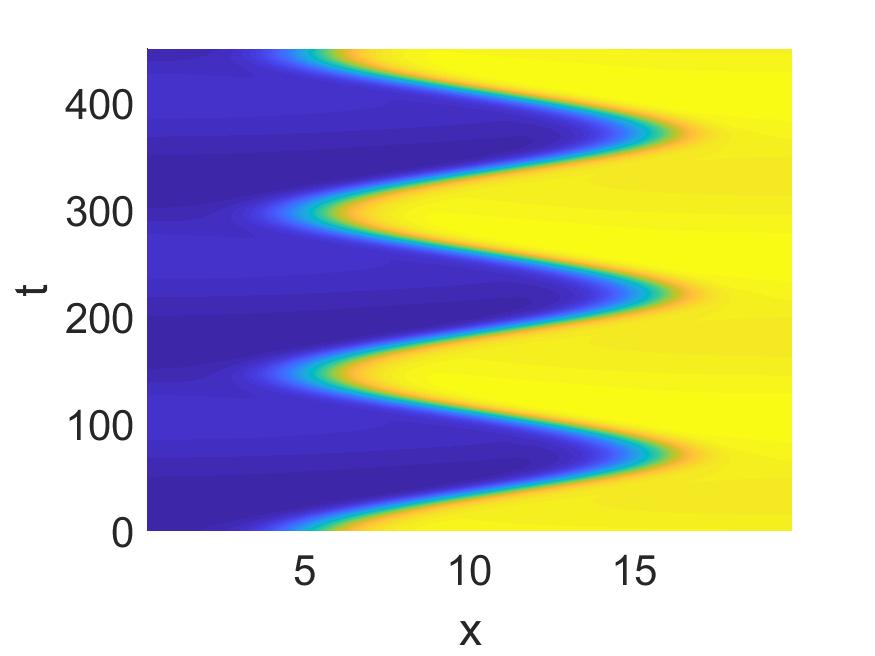}
  }  
     \subfigure[~Gaussian processes]{
  \includegraphics[scale=0.125]{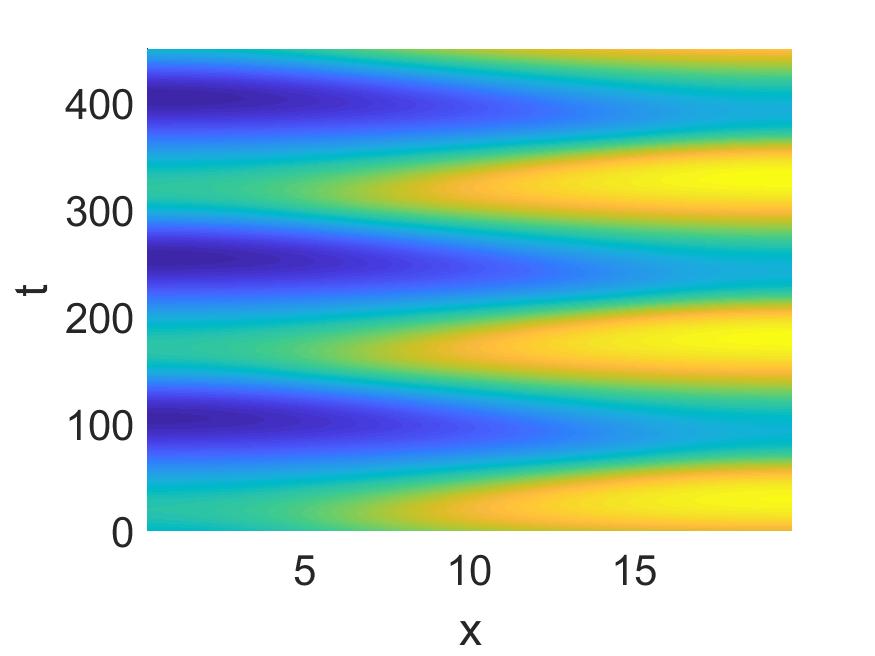}
  } 
      \subfigure[~Neural networks]{
  \includegraphics[scale=0.125]{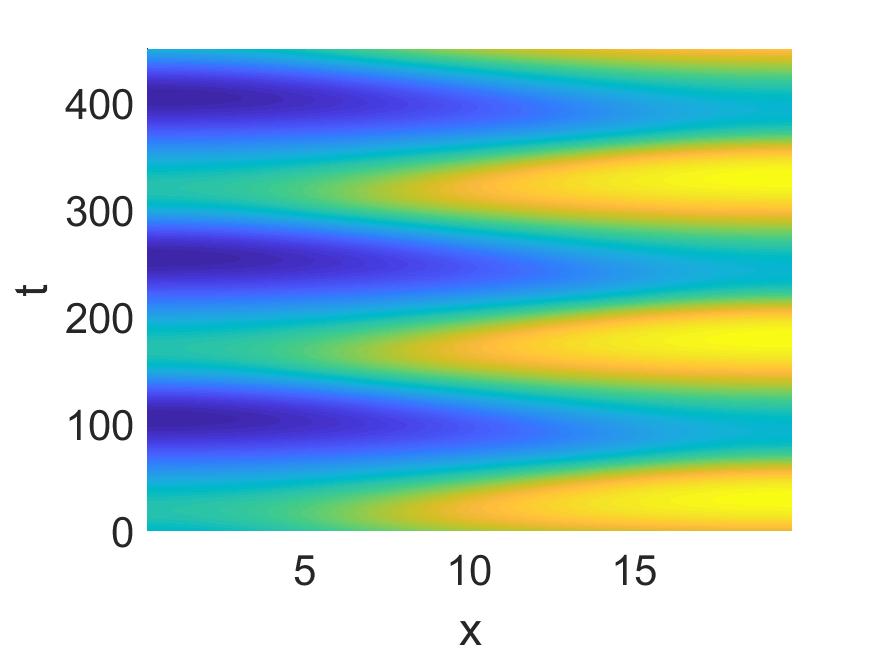}
  }
    \subfigure[~Absolute difference for $u$ (GP)]{
  \includegraphics[scale=0.125]{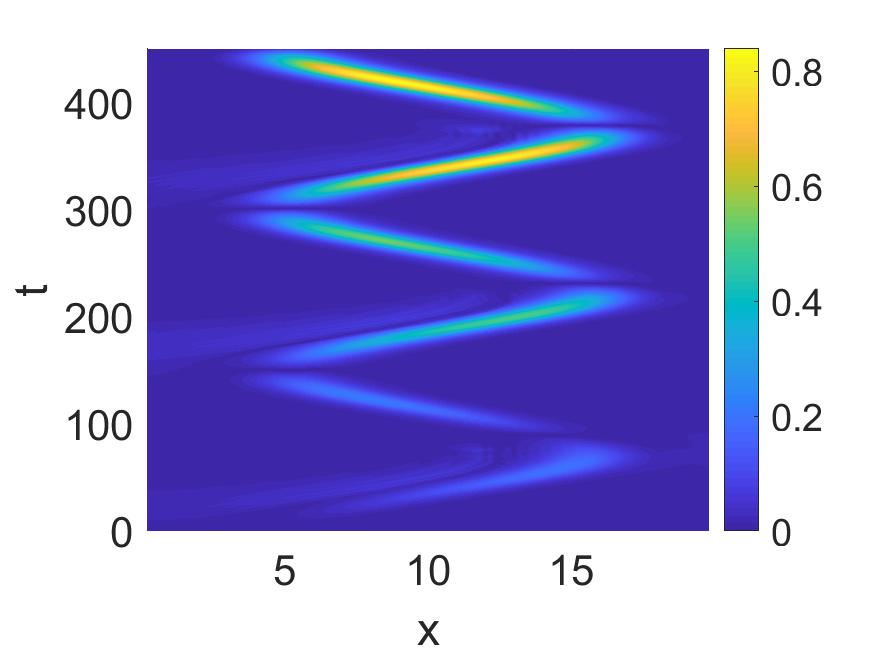}
  } 
    \subfigure[~Absolute difference for $u$ (NN)]{
  \includegraphics[scale=0.125]{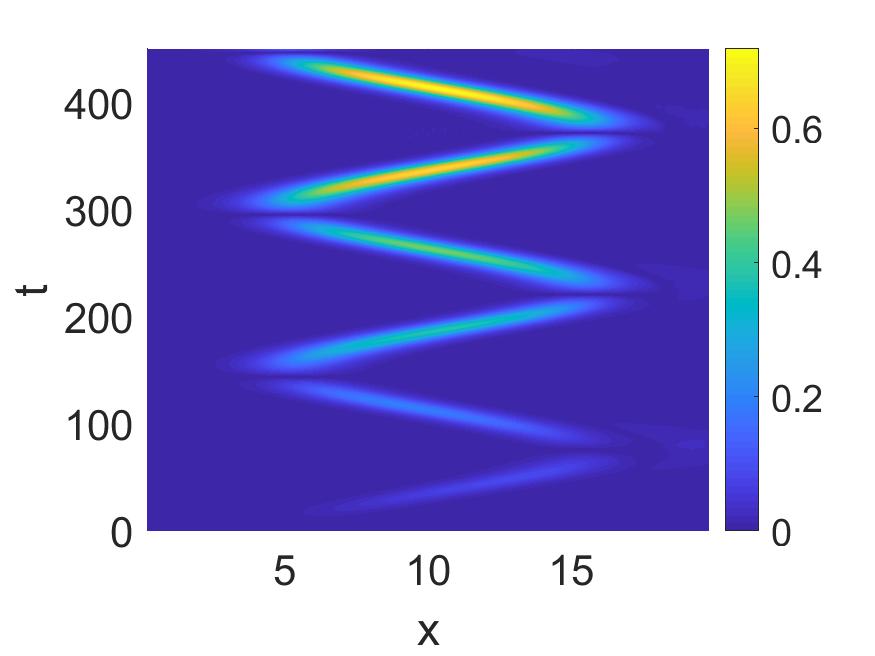}
  }
    \subfigure[~Absolute difference for $v$ (GP)]{
  \includegraphics[scale=0.125]{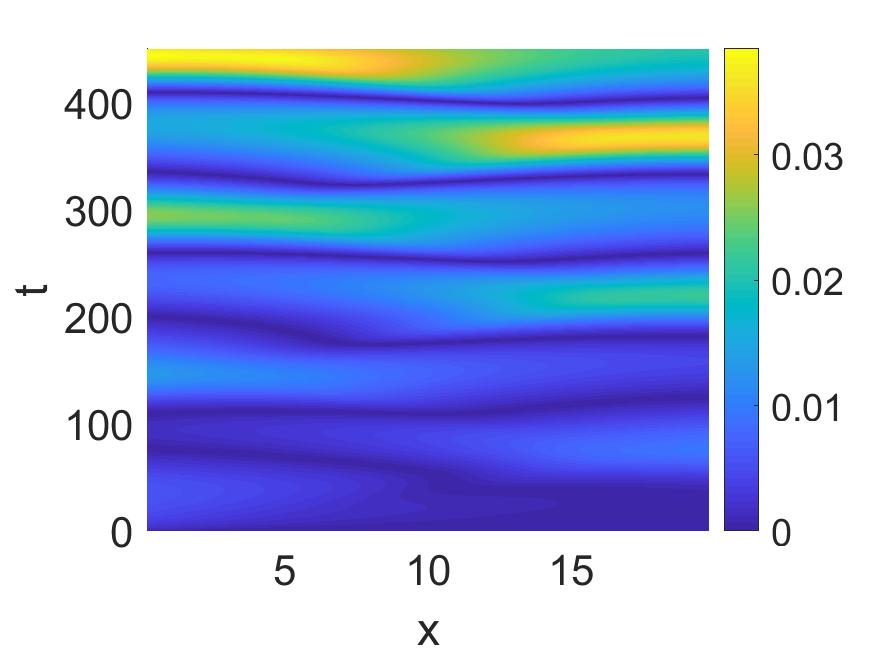}
  } 
    \subfigure[~Absolute difference for $v$ (NN)]{
  \includegraphics[scale=0.125]{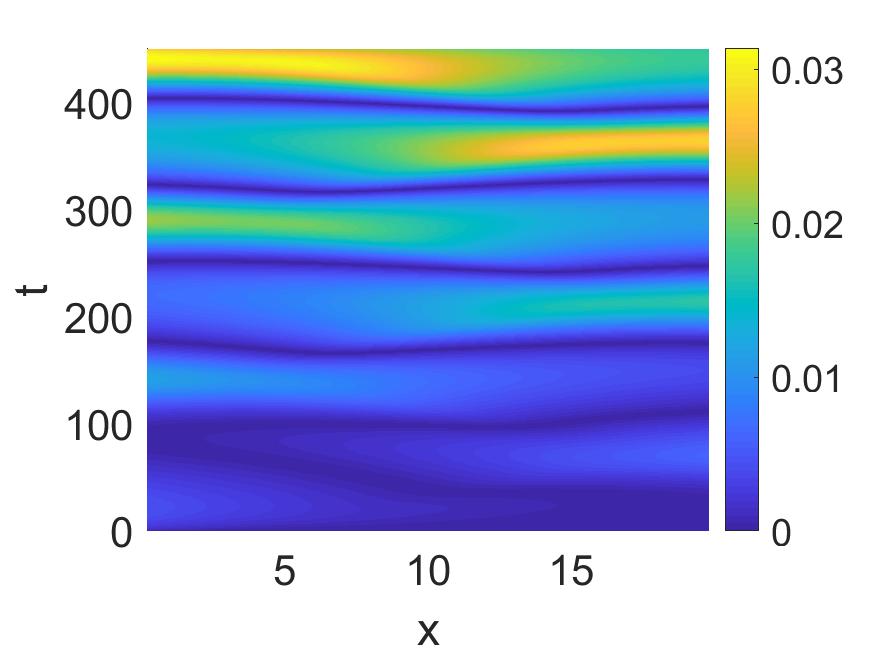}
  }  
  \caption{Feature selection 2: $u_t=f^u(u,u_{x},v)$ and $v_t=f^v(u,v,v_{xx})$.
 (a)-(d): Simulation results of the two methods for $u$ and $v$. (e)-(h): The normalized absolute differences from the ``ground truth" LB simulations for $u$ and $v$.
  }
  \label{fig:f2}
\end{figure}

Using these alternative candidate feature sets, we model different ``versions" of what, on the data, is effectively the same coarse-scale PDE.
The ``alternative" version of the PDE can be symbolically written as
\begin{equation}
\label{eqn:f2}
\begin{aligned}
 u_t(x,t) &= f_2^u(u, u_{x},v),\\
 v_t(x,t) &= f_2^v(u,v, v_{xx}),
\end{aligned}
\end{equation}
and the corresponding regression results of the time derivatives are shown in figure~\ref{fig:derivative_f2}.
Specifically, we use the first spatial derivative $u_x$ instead of the second spatial derivative $u_{xx}$ for learning $u_t$.

As shown in figure~\ref{fig:f2}, both models provide good predictions of the ``ground truth" LB simulations;
we observe, however, that the accuracy of the neural network based predictions is enhanced.
These results confirm that, on the data, alternative coarse-scale PDE forms can provide successful macroscopic description.

\begin{figure}[!htp]
  \centering
   \subfigure[~$u_t$ by GP]{
  \includegraphics[scale=0.17]{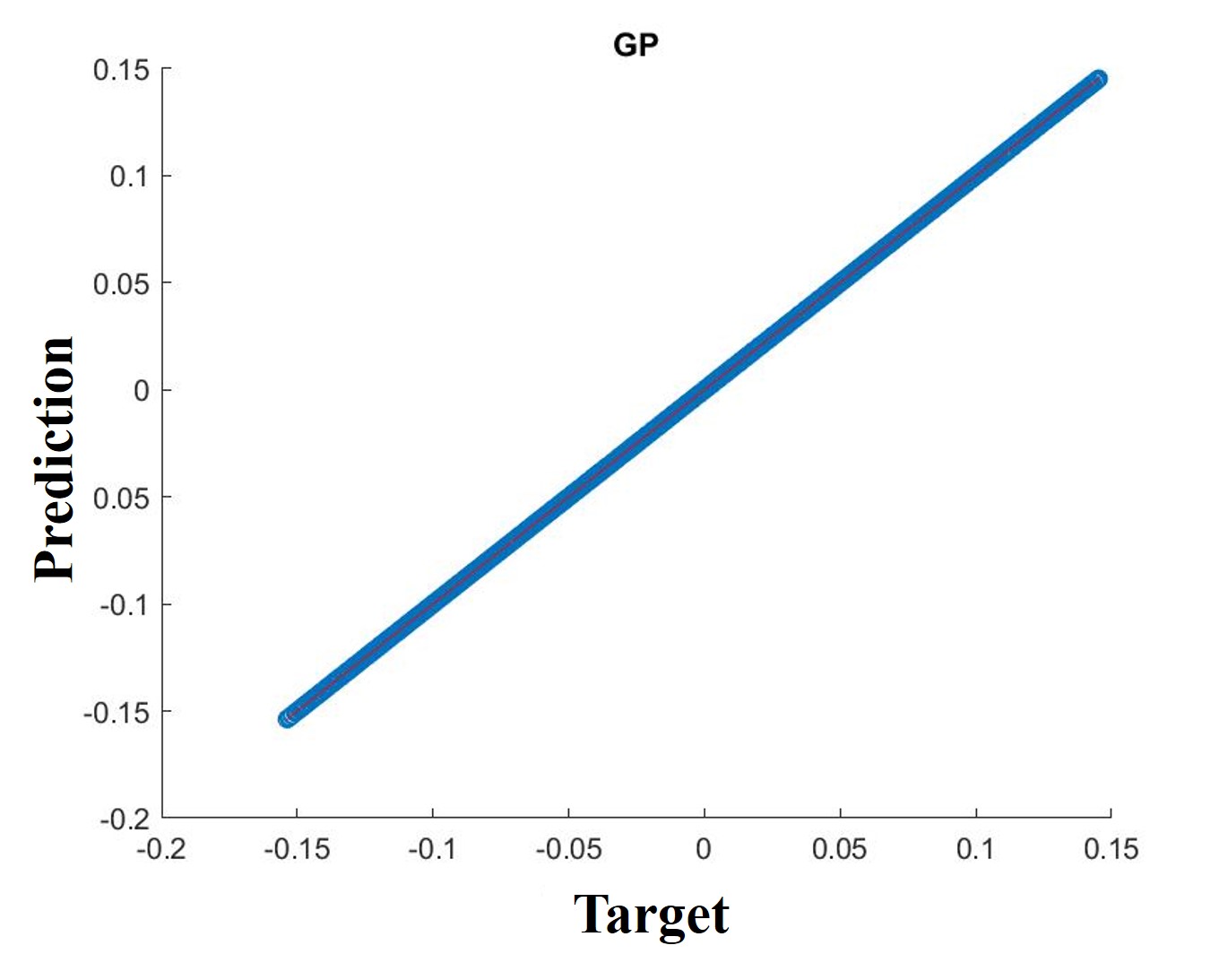}
  }
     \subfigure[~$v_t$ by GP]{
  \includegraphics[scale=0.17]{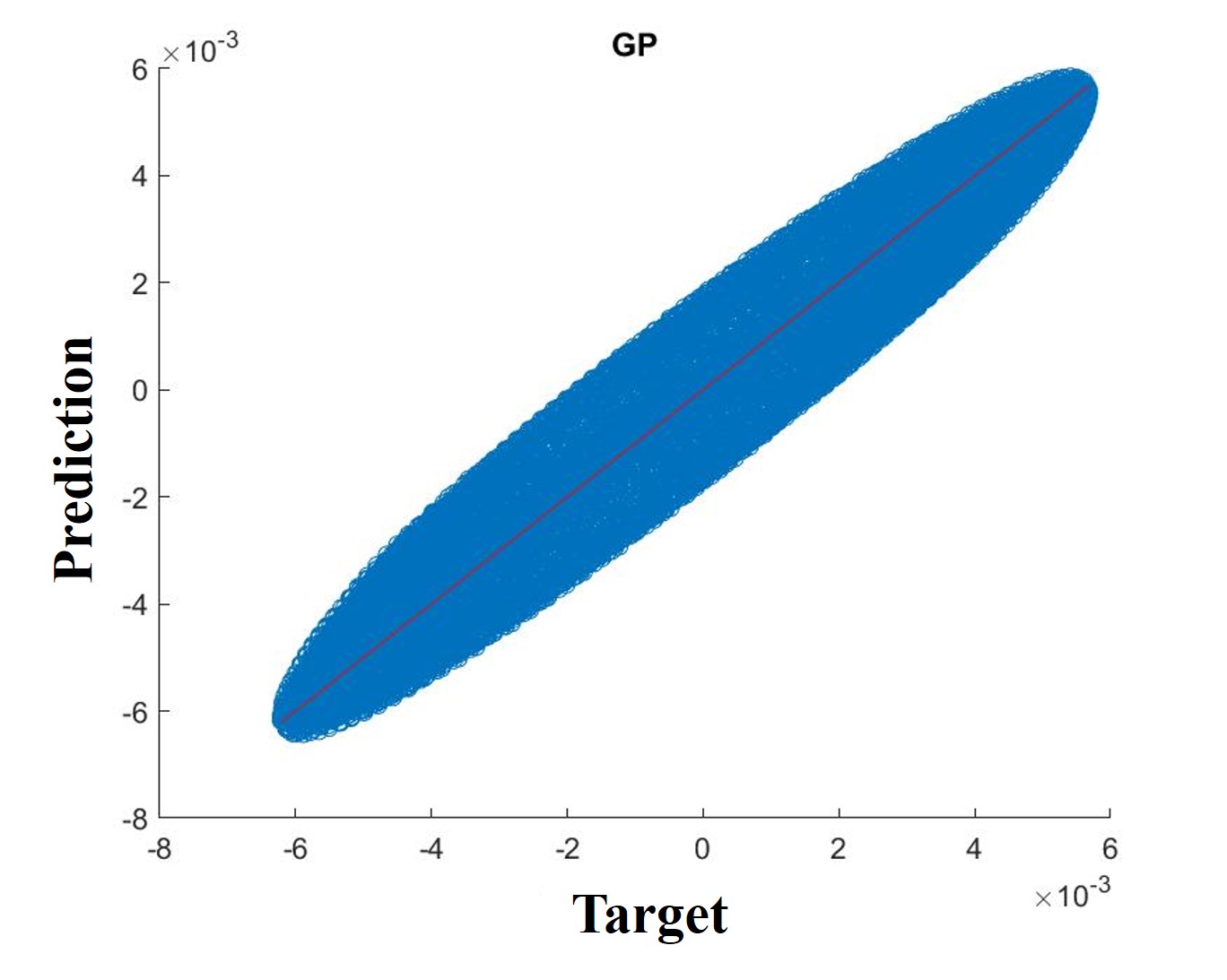}
  }
     \subfigure[~$u_t$ by NN]{
  \includegraphics[scale=0.17]{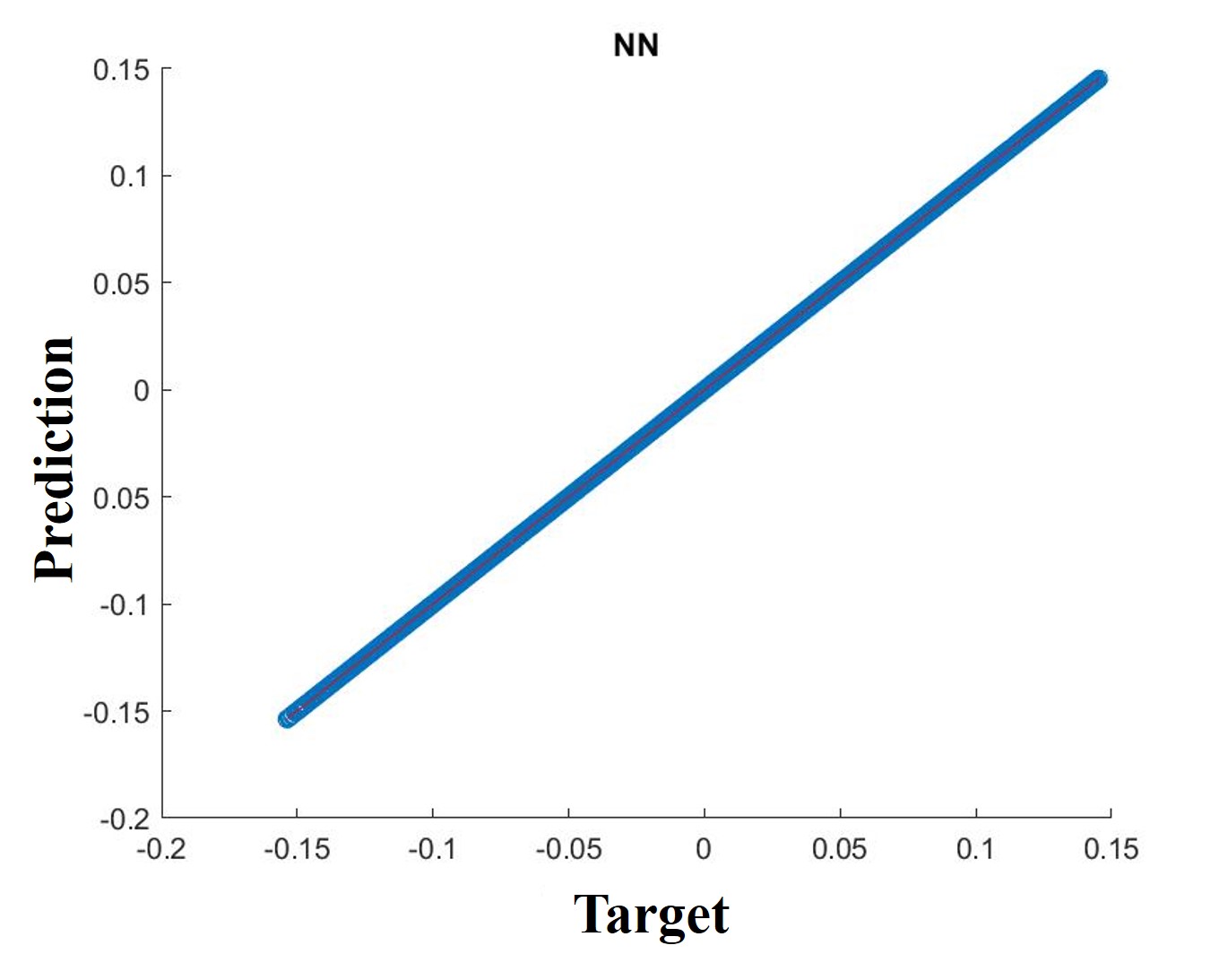}
  }
     \subfigure[~$v_t$ by NN]{
  \includegraphics[scale=0.17]{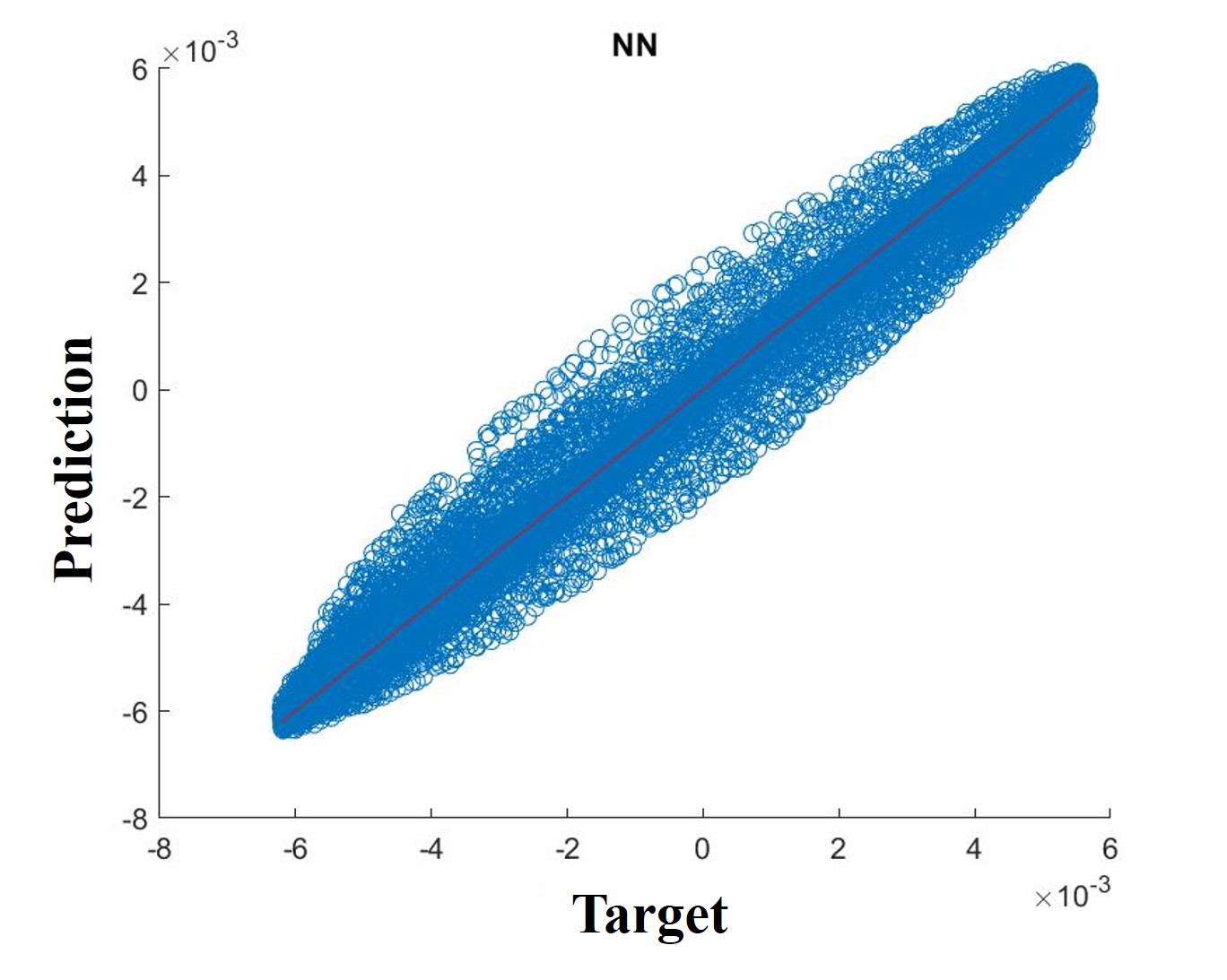}
  }
  \caption{Feature selection 3: $u_t=f_3^u(u,u_{xx},v)$ and $v_t=f_3^v(u,u_{x},v_{xx})$.
  Regression results of the two methods for time derivatives: Gaussian processes (GP) and neural networks (NN).
  }
  \label{fig:derivative_f3}
\end{figure}

To further explore this possibility of alternative PDE forms that represent the observed data with {\em qualitatively comparable accuracy}, we also explored the efficacy of a third coarse-scale PDE description, in terms of a yet different input feature set:
$(u, u_{xx},v)$ for $u_t$ and $(u,u_{x}, v_{xx})$ for $v_t$, so that the PDE can symbolically be written as
\begin{equation}
\label{eqn:f3}
\begin{aligned}
 u_t(x,t) &= f_3^u(u, u_{xx},v),\\
 v_t(x,t) &= f_3^v(u,u_{x}, v_{xx}).
\end{aligned}
\end{equation}
The corresponding prediction results of the time derivatives are shown in figure~\ref{fig:derivative_f3}.
%

\begin{figure}[!htp]
  \centering
    \subfigure[~Gaussian processes]{
  \includegraphics[scale=0.125]{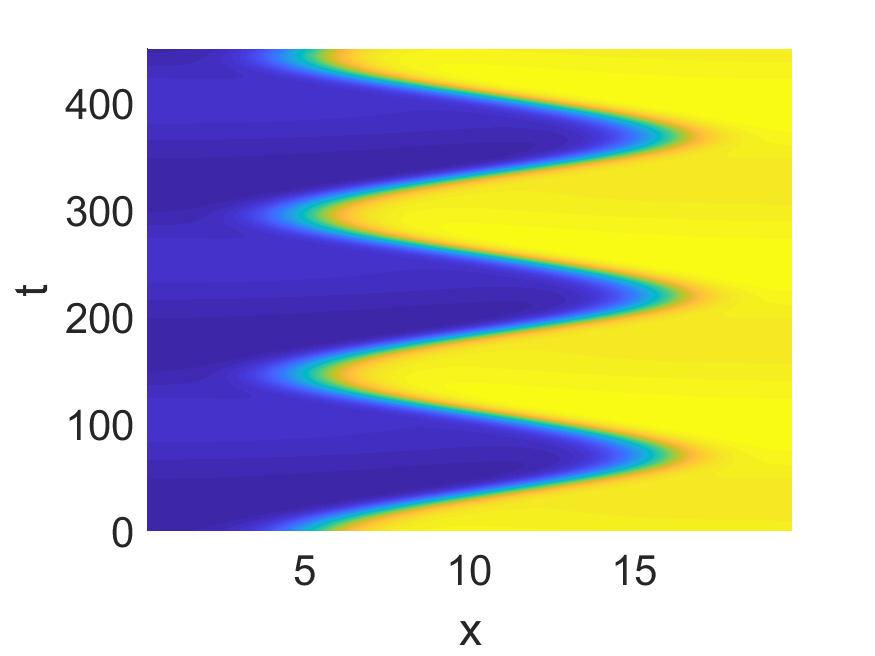}
  }
    \subfigure[~Neural networks]{
  \includegraphics[scale=0.125]{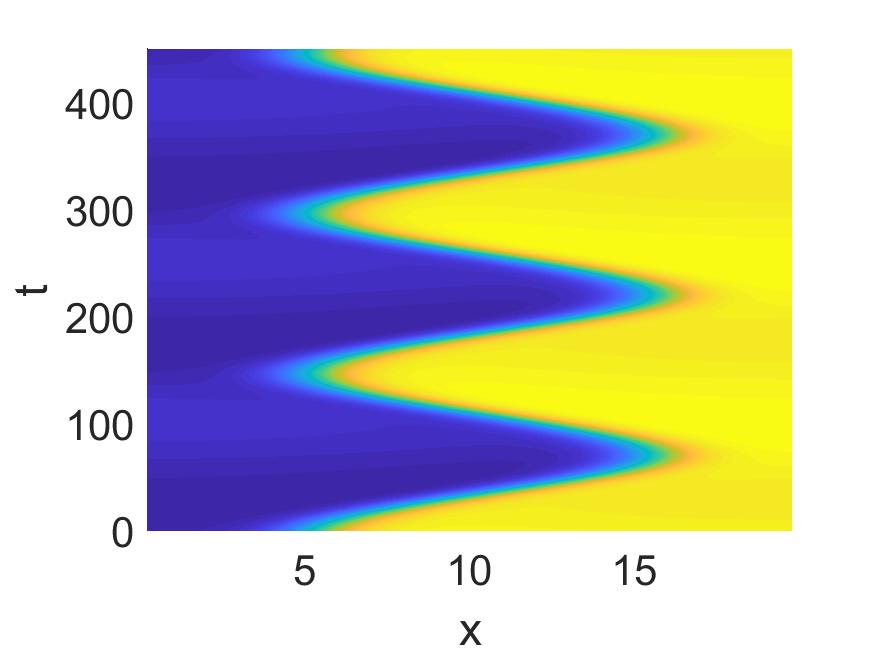}
  }  
     \subfigure[~Gaussian processes]{
  \includegraphics[scale=0.125]{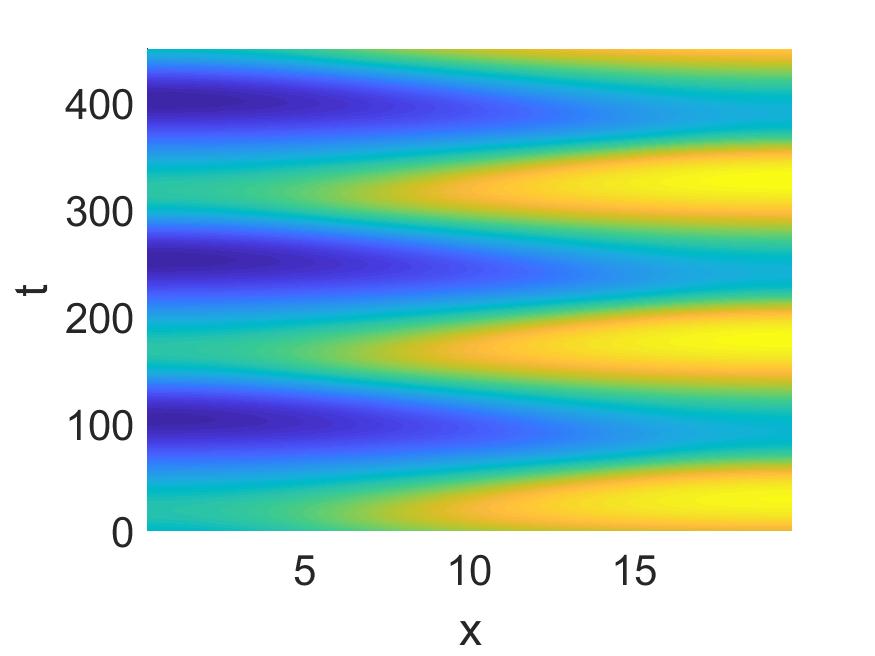}
  } 
      \subfigure[~Neural networks]{
  \includegraphics[scale=0.125]{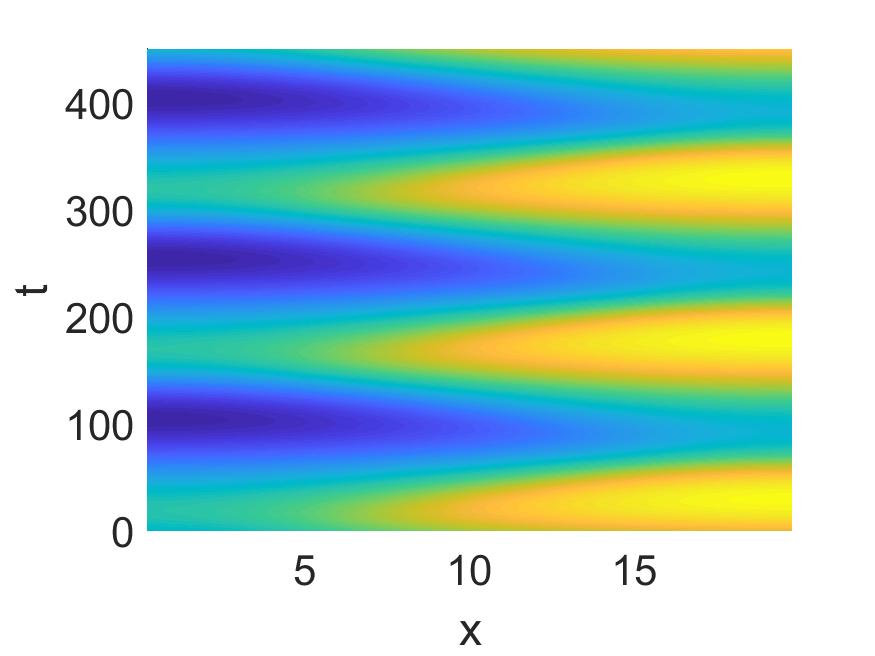}
  }
    \subfigure[~Absolute difference for $u$ (GP)]{
  \includegraphics[scale=0.125]{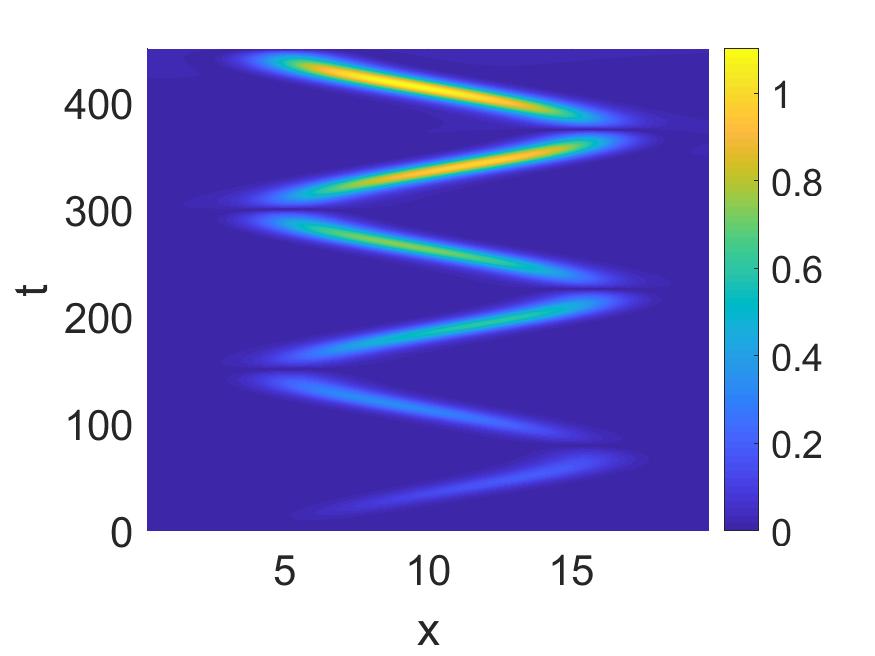}
  } 
    \subfigure[~Absolute difference for $u$ (NN)]{
  \includegraphics[scale=0.125]{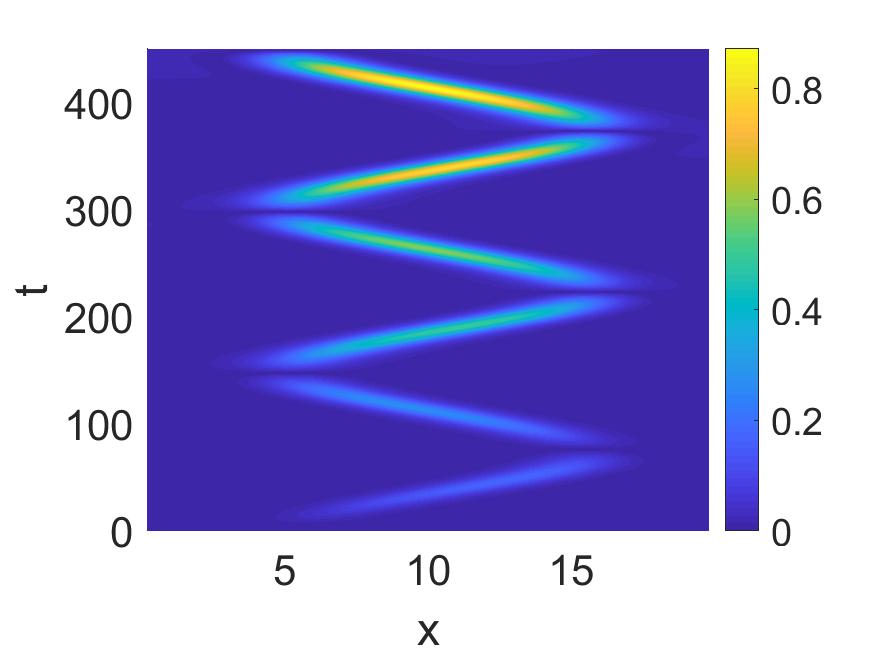}
  }
    \subfigure[~Absolute difference for $v$ (GP)]{
  \includegraphics[scale=0.125]{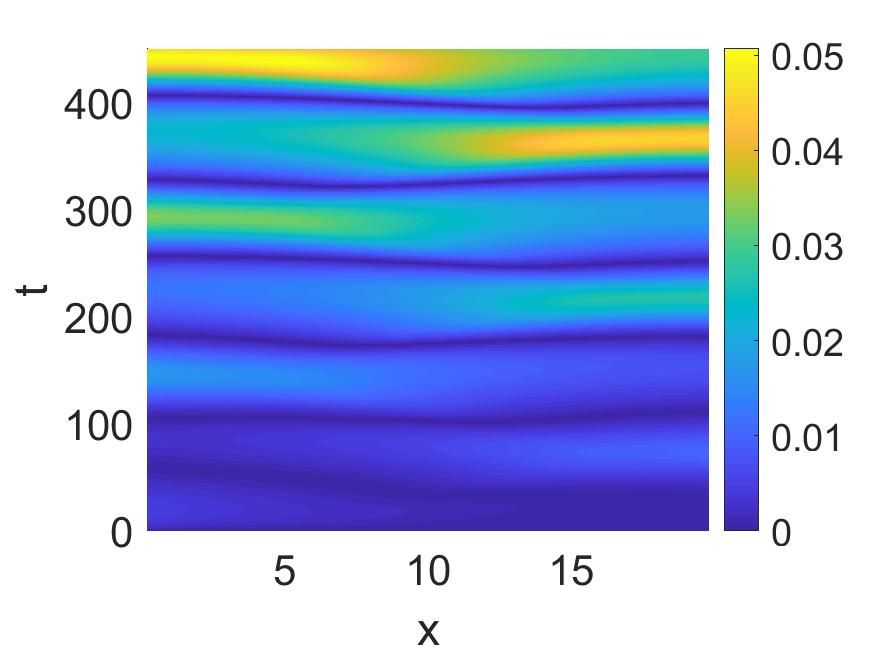}
  } 
    \subfigure[~Absolute difference for $v$ (NN)]{
  \includegraphics[scale=0.125]{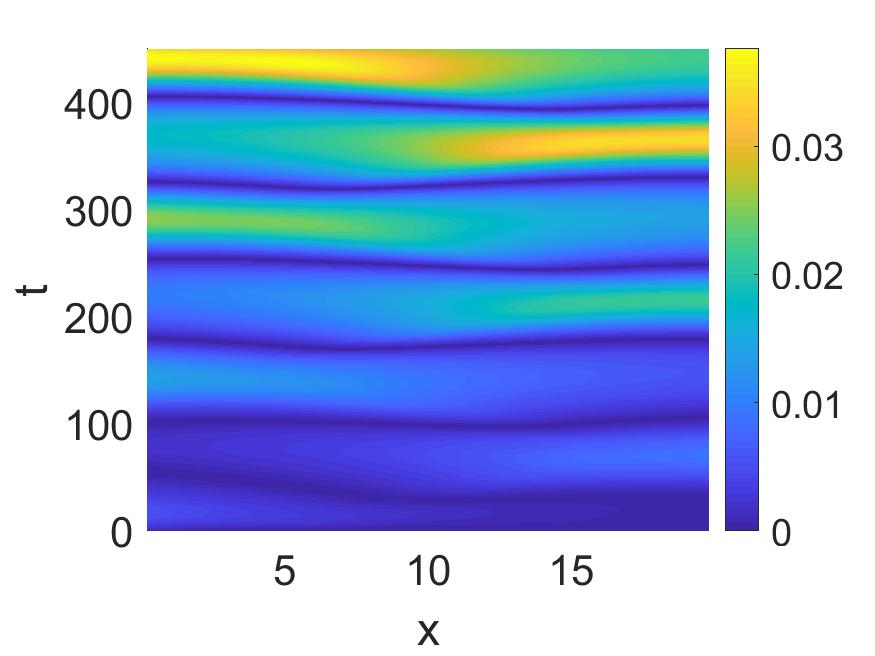}
  }  
  \caption{Feature selection 3: $u_t=f^u(u,u_{xx},v)$ and $v_t=f^v(u,u_{x},v_{xx})$.
  (a)-(d): Simulation results of the two methods for $u$ and $v$. (e)-(h): The normalized absolute differences from the ``ground truth" LB simulations for $u$ and $v$.
  }
  \label{fig:f3}
\end{figure}

As shown in figure~\ref{fig:derivative_f3}, both regression methods provide an inaccurate approximation of $v_t$ near $v_t=0$; the order of magnitude of this error is $10^{-3}$.
The long term prediction results are not as accurate representations of the ground truth LB simulation as the previous two coarse-scale PDE realizations; yet they may still be qualitatively informative.
Normalized absolute differences of long-time simulation for both machine learning methods are shown in figure~\ref{fig:f3}.
As was the case in the previous alternative PDE realizations, the NN model appears more accurate than the GP one.

To compare our identified coarse-scale PDEs with the explicitly known FHN PDE (see equations~\eqref{eqn:fhn}), we also compare the predictions of our coarse-scale PDEs to those of the FHN PDE via mean normalized absolute differences for the test coarse initial condition followed from $t=0$ to $t=450$ as 
\begin{equation}
\label{eqn:mnae}
\begin{aligned}
 \mathrm{MNAD_u} &= \frac{1}{N_T}\sum_{i=1}^{99}\sum_{j=0}^{450} \frac{|u(i,j)-u_f(i,j)|}{\max(u_f)-\min(u_f)},\\
 \mathrm{MNAD_v} &= \frac{1}{N_T}\sum_{i=1}^{99}\sum_{j=0}^{450} \frac{|v(i,j)-v_f(i,j)|}{\max(v_f)-\min(v_f)},
\end{aligned}
\end{equation}
where $N_T$ is a total number of data points and $u_f$ and $v_f$ represent simulation results of the FHN PDE, respectively.
The comparison of these representative simulation of our various coarse-scale PDEs is summarized in table~\ref{tab:mae}.
The differences across our various coarse-scale identified PDEs are of  order $10^{-2}$ and below, comparable to the difference between each of them and the FHN PDE.
\begin{table}[!htp]
\caption{Mean normalized absolute difference (MNAD) for different coarse-scale PDEs. `GP' and `NN' represent `Gaussian processes' and `Neural networks', respectively.}  \label{tab:mae}
\begin{ruledtabular}
\begin{tabular}{lcc}
    & $\mathrm{MNAD}_u$ & $\mathrm{MNAD}_v$\\ \hline
    No Feature selection with GP  & 1.59E-02 & 1.62E-02 \\
    No Feature selection with NN  & 1.53E-02 & 1.56E-02 \\
    Feature selection 1 with GP  & 1.58E-02 & 1.62E-02 \\
    Feature selection 1 with NN  & 1.54E-02 & 1.57E-02 \\
    Feature selection 2 with GP  & 2.39E-02 & 2.20E-02 \\
    Feature selection 2 with NN  & 2.00E-02 & 2.11E-02 \\
    Feature selection 3 with GP  & 3.20E-02 & 3.31E-02 \\
    Feature selection 3 with NN  & 2.08E-02 & 2.16E-02
\end{tabular}
\end{ruledtabular}
\end{table}
Specifically, `feature selection 1' (figure~\ref{fig:f1}), whose variables are the same as those of the explicit FHN PDE, provides the best PDE realization via {\em both} the GP and the NN models.
\section{Conclusion}
\label{sec:conclusion}
In this paper, we demonstrated the data-driven discovery of macroscopic, concentration-level PDEs for reaction/transport processes resulting from fine-scale observations (here, from simulations of a Lattice Boltzmann mesoscopic model).
Long-term macroscopic prediction is then obtained by simulation of the identified (via machine-learning methods) coarse-scale PDE.
We explored the effect of input feature selection capability on the effectiveness of our framework to identify the underlying macroscopic PDEs.

Our framework suggests four different PDEs (one without and three with feature selection), all comparable with the explicit FitzHugh-Nagumo PDE {\em on the data}: all of them provide good approximations of sample coarse-scale dynamic trajectories.
The FHN PDE terms have a well-established mechanistic physical meaning (reaction and diffusion); it would be interesting to explore if any physical meaning can be ascribed to our alternative parametrizations of the right-hand-side of the coarse-scale evolution PDE.
Clearly, the identified PDE depends on our choice of observables - for example, our Diffusion Map embedding coordinates. 
We plan to  explore the use of kernel-based embeddings  (as discussed in Ref.~\cite{Bittracher19} mentioned above) as an approach that can control the well-posedness of the embedding and the distortion of the resulting manifold; this will clearly affect the identified PDE, and it will be interesting to study the interplay between differently distorted embedding manifolds and different identified approximate PDEs. 

In our framework, we employed finite differences to estimate spatial derivatives in the formulation of the PDE.
Instead of numerical spatial derivatives, we may use the values of coarse observables at neighboring points directly to uncover the coarse evolution law. 
The effect of this alternative embedding for the PDE right-hand-side, explored in Ref.~\cite{Bar19}, on the accuracy of the identified model predictions, is the subject of ongoing research.

We believe that the framework we presented is easily generalizable to multiscale and/or multifidelity data.
Here we worked across a single scale gap and a single fine-scale simulator providing the ground truth. 
We envision that data fusion tools can be combined with the approach to exploit data at several cascaded scales, and taking advantage of simulation data from  several heterogeneous simulators~\cite{Lee17D,Lee19}.
\begin{acknowledgments}
S. Lee, M. Kooshkbaghi, and I. G. Kevrekidis gratefully acknowledge partial support by NIH and by DARPA. Also, This material is based upon work supported in part by, the U. S. Army Research Laboratory and the U. S. Army Research Office under contract/grant number W911NF1710306.
Discussions of the authors with Dr. Felix Dietrich are gratefully acknowledged.
\end{acknowledgments}
\appendix
\section{``Healing" for the Lattice Boltzmann model}
\label{sec:heal}
An important assumption underlying our work is that the fine-scale model can ``close" at a coarse-scale level.
In our particular case, this means that, even though our fine scale Lattice Boltzmann model (LBM) evolves particle distribution functions, one can be predictive at the coarse level of the zeroth moments of these functions, the concentrations $u$ and $v$ of the activator and the inhibitor.
The hypothesis that allows this reduction is that the problem is {\em singularly perturbed} in time: higher order moments of these distribution functions become quickly slaved to the ``slow", governing, zeroth order moment fields.
Yet, while initializing the FHN PDE only requires spatial profiles of $u$ and $v$ at the initial time, initializing the full FHN LBM requires initial conditions for {\em all} the evolving particle distributions.
Constructing such detailed fine scale initializations consistent with the coarse initial conditions is an important conceptual component of equation-free computation; the term used is ``lifting" 
\cite{Kevrekidis03,Kevrekidis09}.
%
\begin{figure}[!htp]
  \centering
  \includegraphics[width=0.45\textwidth]{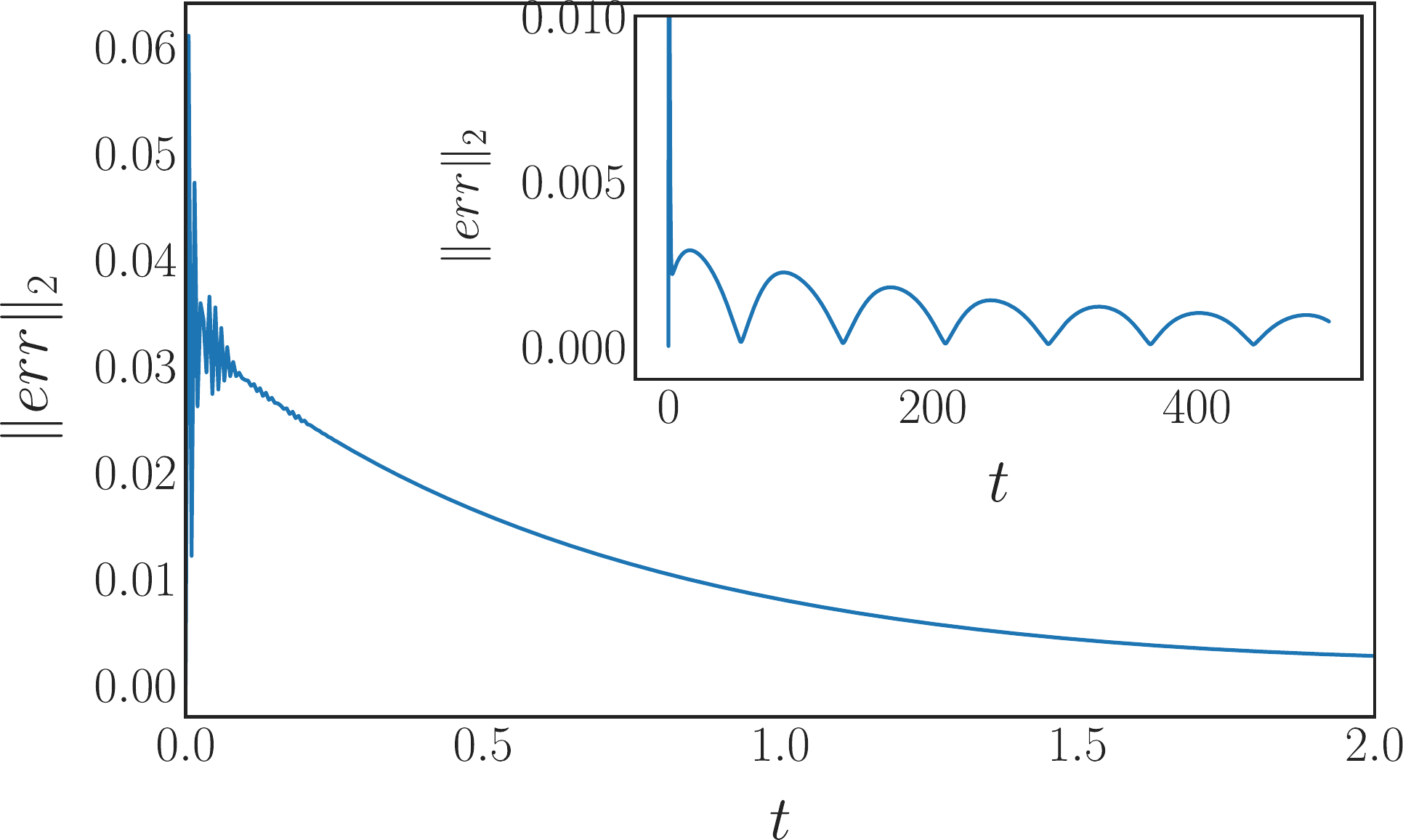}
  \caption{Evolution of the $L^2$ norm (see equation~\eqref{eq:l2norm_diff}) of the coarse difference between trajectories with the same coarse but different fine initial conditions. 
  After the initial small (but violent) oscillation in error abates (for $t \lesssim 0.1$), the perturbed system relaxes to the vicinity of the base solution over $t \approx 2$.}
  \label{fig:relax}
\end{figure}

Here, in lifting the coarse-scale observable (a concentration field $\rho$) to the microscopic description (particle distribution function $f$) on each node, we employ an equal weight rule, i.e., the three particle distribution function values at the node $x_n$ are chosen to be
\begin{equation}
    f_{-1}(x_n)=f_{0}(x_n)=f_{1}(x_n) = \frac{\rho(x_n)}{3}.
\end{equation}
This equal weight choice (the local, spatially uniform diffusive equilibrium distribution) is not, in general, consistent with the (spatially nonuniform, and not simply diffusive) macroscopic PDE model (here, the FitzHugh-Nagumo PDEs); yet we expect that the fine scale simulation features will become rapidly slaved to the local concentration field~\cite{Van05}. 
To estimate the appropriate slaving/relaxation time,
we compare the $L^2$ norm of the density predicted by two differently initialized LBM simulations: one that lies on the long-term stable limit cycle, and one that results from it by retaining the coarse state, but perturbing the associated fine scale states according to the local diffusive equilibrium $\frac{1}{3}$ rule (for $\epsilon=0.01$ in equation~\eqref{eqn:reaction}).

To explore the slaving time scale, we trace the $L^2$ norm of the difference between the simulations resulting from these two initializations.
This $L^2$ norm is defined as 
\begin{equation}
\Vert err \Vert_2 = \Vert \rho_{eq}(x,t)- \rho(x,t)\Vert_2,
\label{eq:l2norm_diff}
\end{equation}
where $\rho_{eq}$ and $\rho$ represent the density with equal weights and the density without equal weights (reference solution), respectively.
As shown in figure~\ref{fig:relax}, after a fast transient oscillation of $L^2$ (for $t < 0.1$), the 
norm decays smoothly until $t \approx 2$.
There is still a small inherent bias (the trajectory will come back to a {\em nearby} point along the limit cycle); this does not affect our estimate of the slaving time. 
We therefore chose a relaxation time $t=2$ (or 2000 LB time steps) and we started collecting coarse observations as training data from our various initializations only after $t=2$. 
%
\nocite{*}
\bibliography{aareferences}

\end{document}